\documentclass[10pt,twocolumn,letterpaper]{article}

\usepackage{cvpr}
\usepackage{times}
\usepackage{epsfig}
\usepackage{graphicx}
\usepackage{amsmath}
\usepackage{amssymb}
\usepackage{float}
\usepackage{relsize}
\usepackage{subcaption}

\usepackage{algpseudocode}
\usepackage{algorithm,}

\cvprfinalcopy 

\def\cvprPaperID{3466} 

\begin{document}

\title{Moving Object Detection under Discontinuous Change in Illumination Using Tensor Low-Rank and Invariant Sparse Decomposition}

\author{Moein Shakeri\\
University of Alberta\\
Edmonton, AB, Canada\\
{\tt\small shakeri@ualberta.ca}
\and
Hong Zhang\\
University of Alberta\\
Edmonton, AB, Canada\\
{\tt\small hzhang@ualberta.ca}
}

\maketitle

\begin{abstract}
   Although low-rank and sparse decomposition based methods have been successfully applied to the problem of moving object detection using structured sparsity-inducing norms,~they~are~still~vulnerable to significant illumination changes that arise in certain applications.~We~are~interested in moving object detection in applications involving time-lapse image sequences for which current methods mistakenly group moving objects and illumination changes into foreground.~Our method relies on the multilinear (tensor) data low-rank and sparse decomposition framework to address the weaknesses of existing methods.~The key to our proposed method is to create first a set of prior maps that can characterize the changes in the image sequence due to illumination.~We~show that they can be detected by a $k$-support norm.~To deal with concurrent, two types of changes, we employ two regularization terms, one for detecting moving objects and the other for accounting for illumination changes, in the tensor low-rank and sparse decomposition formulation.~Through comprehensive experiments using challenging datasets, we show that our method demonstrates a remarkable ability to detect moving objects under~discontinuous~change~in~illumination,~and outperforms the state-of-the-art solutions to this challenging problem.
\end{abstract}

\section{Introduction}

\begin{figure}[t]
\centering
\includegraphics[width=\linewidth, height=3.5cm]{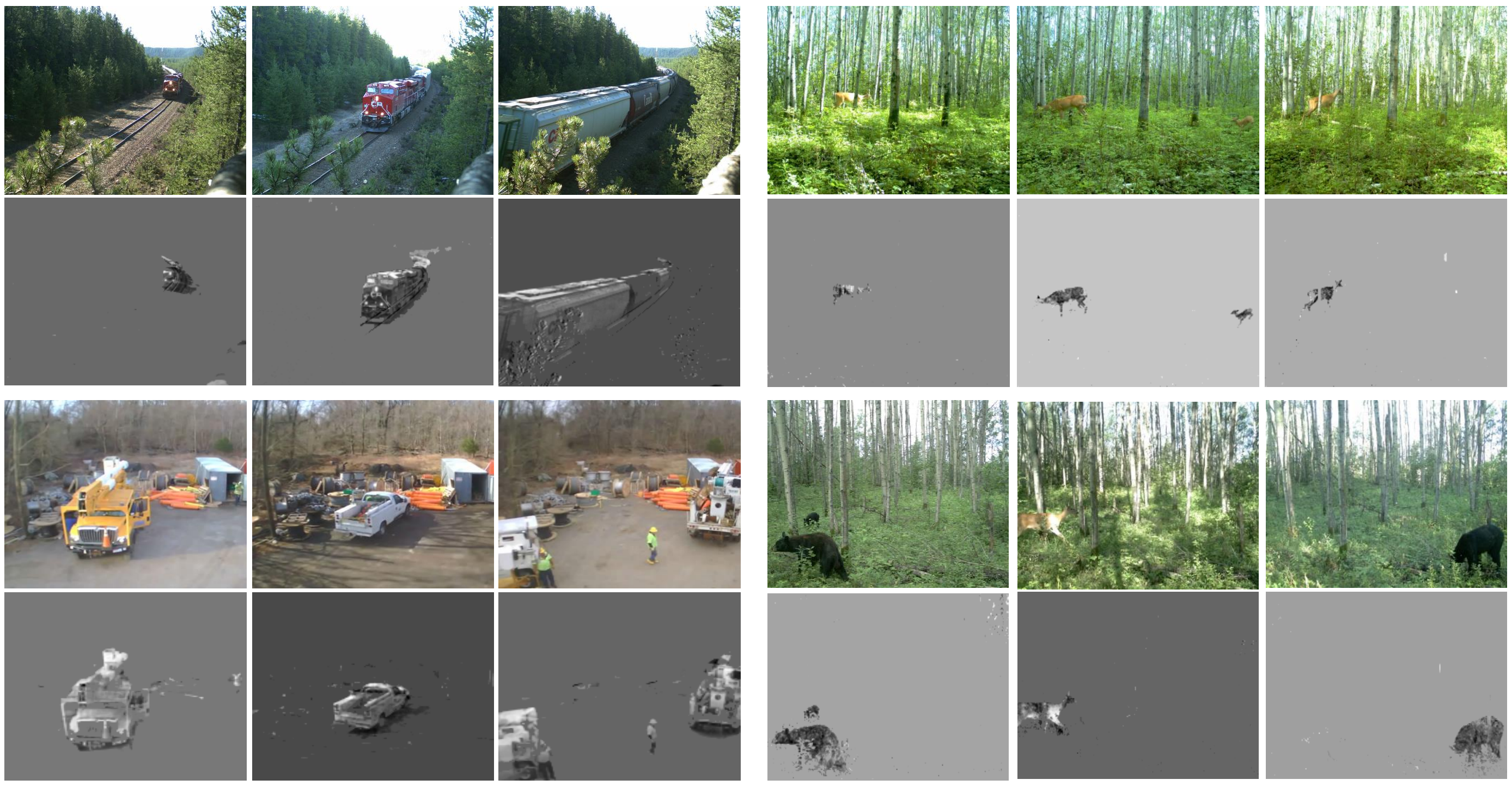}
\caption{First row of each sequence: images captured in a industrial or wildlife monitoring system. Second row: results of our proposed method to detect foreground objects.}
\label{sample_problem}
\vspace{-15pt}
\end{figure}

Moving object detection in an image sequence captured under uncontrolled illumination conditions is a common problem in computer vision applications such as visual surveillance~\cite{int_video_surveillance}, traffic monitoring~\cite{int_traffic_monitoring}, and social signal processing~\cite{int_object_encoding}. Although moving object detection and background subtraction is a well established area of research and many solutions have been proposed, still most of the existing solutions are vulnerable to complex illumination changes that frequently occur in practical situations, especially when the changes are discontinuous in time. In such cases, current methods are often not able to distinguish between illumination changes (including those due to shadow), and changes caused by moving objects in the scene. In general, outdoor illumination conditions are uncontrolled, making moving object detection a difficult and challenging problem. This is a common problem for many surveillance systems in industrial or wildlife monitoring areas in which a motion triggered camera or a time-lapse photography system is employed for detecting objects of interest over time. Fig.~\ref{sample_problem} shows four image sequences under discontinuous changes in illumination, which illustrate these applications. Due to significant and complex changes in illumination and independent changes of the moving objects between images of the sequences, detection of the moving objects is extremely challenging. The second row of each sequence in Fig.~\ref{sample_problem} shows the sample results of our proposed method with detected moving objects.

Among the leading methods for the problem addressed in this paper is a group based on low-rank and sparse decomposition. This group of methods exploit the fact that the background in an image sequence can be described as a low-rank matrix whose columns are image pixels that are correlated~\cite{int_low_01,int_low_02}.
However, image sequences with moving objects under discontinuous change in illumination and object location using the timer-lapse photography are qualitatively different from regular frame-rate video sequences. While some existing solutions are able to handle the discontinuity in object location with limited success, there is a need to improve their ability to distinguish between moving objects and changes due to illumination.

Taking the idea of using the low-rank components of a matrix to capture the image background, the most recent development relies on tensors, which are higher dimensional data structures than $2D$-matrices. Since the real world data are ubiquitously multi-dimensional, tensors are often more appropriate than $2D$-matrices to capture higher order relations in data. It is not surprising that tensor low-rank methods have been successfully developed with promising results on real-time video sequences. However, such methods have yet to be studied for detecting moving objects under discontinuous changes in illumination and object position, such as those found in time-lapse image sequences.

In this paper, we propose a solution to the problem of moving object detection within the tensor low-rank framework that specifically addresses the problem of discontinuous changes in illumination and object location. We formulate the problem in a unified framework named tensor low-rank and invariant sparse decomposition (TLISD). To separate illumination changes from moving objects, first we compute multiple prior maps as illumination invariant representations of each image to build our tensor data structure. These prior maps provide us with information about the effect of illumination in different parts of an image. We show that by defining two specific penalty terms using these prior maps, our proposed method is able to decompose an image into background, illumination changes and foreground objects, with a significant boost in performance of moving object detection. 

The main contributions are as follows.
\begin{itemize}
\vspace{-5pt}
\item {We propose to use multiple priors  to model the effect of illumination in natural images by exploiting invariance properties of color image chromaticity.}
\vspace{-6pt}
\item {We make use of the priors in a tensor representation for the problem of moving object detection.}
\vspace{-6pt}
\item {We propose a low-rank tensor decomposition using group sparsity and k-support norm as two regularization terms to separate moving objects and illumination variations that undergo discontinuous changes.}
\vspace{-6pt}
\item {We introduce an extended illumination change dataset with over 80k real images captured by motion trigger cameras in industrial and wildlife monitoring systems.}
\end{itemize}

\section{Related Work}
\label{related}

One successful approach to moving object detection attempts to decompose a matrix $D$ representing an image sequence into a low-rank matrix $L$ and sparse matrix $S$, so as to recover the background and the foreground~\cite{int_low_03}. The problem is initially solved by the robust principal component analysis (RPCA). Since the foreground objects are described by the sparse matrix $S$, we can categorize existing methods by the types of constraints on $S$. The first group of these methods use $l_{1}$-norm to constrain $S$~\cite{int_low_02,int_ssgodec,int_prmf} and solve the following convex optimization.
\vspace{-4pt}
\begin{equation}
\label{eq:1}
\min_{L,S}\|L\|_{*}+\lambda\|S\|_{1} \,\,\,\,s.t.\,\,D=L+S
\vspace{-4pt}
\end{equation}
where $\|L\|_{*}$ denotes the nuclear norm of matrix $L$, and $\|S\|_{1}$ is the $l_{1}$-norm of $S$. 

The second group of methods used the additional prior knowledge on the spatial continuity of objects to constrain sparse matrix $S$ and improve the detection accuracy~\cite{int_low_05,int_low_07}. Using spatial continuity (e.g., $l_{2,1}$-norm in~\cite{int_low_05}) to enforce the block-sparsity of the foreground, results become more stable than conventional RPCA in the presence of illumination changes. However, 
it remains a challenge to handle moving shadows or significant changes in illumination. Furthermore, the position of an object in a time-lapse image sequence is discontinuous from one image to another so that the continuity assumption is invalid as a way to separate moving objects and changes in illumination.

The third group of methods also imposed the connectivity constraint on~$S$~\cite{int_low_06,int_brmf,int_low_08,int_low_09,int_low_091,int_low_10} using other formulations than the second group. For example, 
Liu {\it{et al.}}~\cite{int_low_10} attempted to use a structured sparsity norm~\cite{nips1} and a motion saliency map, to improve the accuracy of moving object segmentation under sudden illumination changes. However, this method still cannot handle shadows and severe illumination changes, especially in time-lapse sequences with independent object locations among the images in the sequence that change similarly to shadow and illumination. In general, although the low-rank framework is well-known to be robust against moderate illumination changes in frame-rate sequences, the existing methods are still not able to handle discontinuous change in illumination and shadow, especially in time-lapse sequences.

To effectively separate discontinuous changes due to moving objects and those due to illumination, Shakeri {\it{et al.}}~\cite{shakeri_ICCV} proposed a method called LISD. This method relies on 
an illumination regularization term combined with the standard low-rank framework to explicitly separate the sparse outliers into sparse foreground objects and illumination changes. Although this regularization term can significantly improve the performance of object detection under significant illumination changes, LISD assumes a) the invariant representation~\cite{shakeri_iros} of all images in a sequence are modeled by only one invariant direction and b) all illumination variations are removed in the invariant representation of images, which are not strictly valid in practice.  

Recently, multi-way or tensor data analysis has attracted much attention and has been successfully used in many applications. Formally and without loss of generality, denote a 3-way tensor by $\mathcal{D} \in R^{n_1 \times n_2 \times n_3}$. Tensor low-rank methods attempt to decompose $\mathcal{D} \in R^{n_1 \times n_2 \times n_3}$ into a low-rank tensor $\mathcal{L}$ and an additional sparse tensor $\mathcal{S}$~\cite{goldfarb_tensor}. This decomposition is applicable in solving many computer vision problems, including moving object detection.
One of the most recent methods relevant to our research is proposed by Lu~{\it{et al.}}~\cite{Lu_tensor}. A tensor nuclear norm was used to estimate the rank of tensor data and RPCA was extended from 2D to 3D to formulate the following tensor robust PCA (TRPCA):
\vspace{-4pt}
\begin{equation}
\label{eq:tensor_1}
\min_{\mathcal{L},\mathcal{S}}\|\mathcal{L}\|_{*}+\lambda\|\mathcal{S}\|_{1} \,\,\,\,s.t.\,\,\mathcal{D}=\mathcal{L}+\mathcal{S}
\vspace{-3pt}
\end{equation}
They showed that the tensor nuclear norm on tensor data can capture higher order relations in data.~Tensor data is used for background subtraction and foreground detection~\cite{sobral_tensor,javed_tensor,tensor_salient_moving, tensor_continue_2016, tensor_2018} by stacking two dimensional images into a three dimensional data structure, using which tensor decomposition can capture moving object due to the continuity of object positions in the third dimension. Obviously, this approach only works for frame-rate sequences with continuous foreground motion, but is not applicable to time-lapse image sequences with discontinuous changes in both object location and illumination.


In this paper, we introduce a new formulation for moving object detection under the framework of tensor low-rank representation and invariant sparse outliers. We first build a set of prior maps for each image in the image sequence and treat it as a tensor. These prior maps enable us to use two regularization terms to distinguish between moving objects and illumination changes. We demonstrate that their use within our proposed method significantly improves the performance of moving object detection in the case of discontinuous changes in illumination, a problem that most of the existing methods cannot handle effectively.
\section{Tensor Low-Rank and Invariant Sparse Decomposition}
\label{proposed}

Our proposed formulation seeks to decompose tensor data $\mathcal{D}$ into a low-rank tensor $\mathcal{L}$, an illumination change tensor $\mathcal{C}$, and a sparse foreground tensor $\mathcal{S}$ as follows.
\vspace{-1pt}
\begin{equation}
\label{eq:ours_1}
\mathcal{D}=\mathcal{L}+\mathcal{S}+\mathcal{C}
\vspace{-1pt}
\end{equation}
In~(\ref{eq:ours_1}), both $\mathcal{S}$ and $\mathcal{C}$ are stochastic in time-lapse image sequences due to discontinuous change in object locations and illumination changes, and separating them is an ill-posed problem. To solve this issue, we compute a set of prior maps using multiple representations of an image, which are more robust against illumination change than RGB images. These prior maps enable us to find higher order relations between the different invariant representations and the intensity images, in both space and time. These relations are exploited as the basis for separating $\mathcal{S}$ from $\mathcal{C}$ as will be detailed in Section~\ref{initial_map}. 
It is worth mentioning that on one hand, illumination changes are related to the material in a scene, which is invariant in all frames leading to a correlation between them. On the other hand, these changes are also related to the source of lighting, which is not necessarily correlated between frames. Consequently, illumination changes should be accounted for by both the low-rank part and the sparse part in an image decomposition. In our method, we model the highly correlated part of illumination with the low-rank tensor $\mathcal{L}$ as background, and we model the independent changes in illumination as the foreground, while recognizing that uncorrelated illumination changes are not necessarily sparse.
To accomplish such illumination modeling, we propose to use a balanced norm or $k-$support norm. We introduce our formulation in details in Section~\ref{sec3_proposed}, and we describe a solution to the formulation in Section~\ref{sec3_optimization}.
\subsection{Generation of Prior Maps and Tensor Data $\mathcal{D}$}
\label{initial_map}

In this section we focus on obtaining the prior information that will enable us to distinguish between moving objects and illumination changes in our proposed formulation. In the case of discontinuous change in illumination, which is common in time-lapse image sequences, variation of shadows and illumination are unstructured phenomena and they are often mistakenly considered by many methods as moving objects. We address this problem through creating illumination-invariant and shadow-free images, a problem that has been well studied. 

One of the most popular methods for this problem is proposed by Finlayson {\it{et al.}}~\cite{IIR_SH}, which 
computes the two-vector log-chromaticity $\chi'$ using red, green and blue channels.~\cite{IIR_SH} showed that with changing illumination, $\chi'$ moves along a straight line $e$ roughly. Projecting the vector $\chi'$ onto the vector orthogonal to $e$, which is called invariant direction, an invariant representation $I=\chi' e^{\perp}$ can be computed. This method works well when the assumption defined above hold true but in practice this assumption never holds exactly, i.e., $\chi'$ does not move along a straight line. As a result, the correspond invariant representation is flawed
and can lead to sub-optimal performance.

\begin{figure}[t!]
\centering
\includegraphics[width=\linewidth]{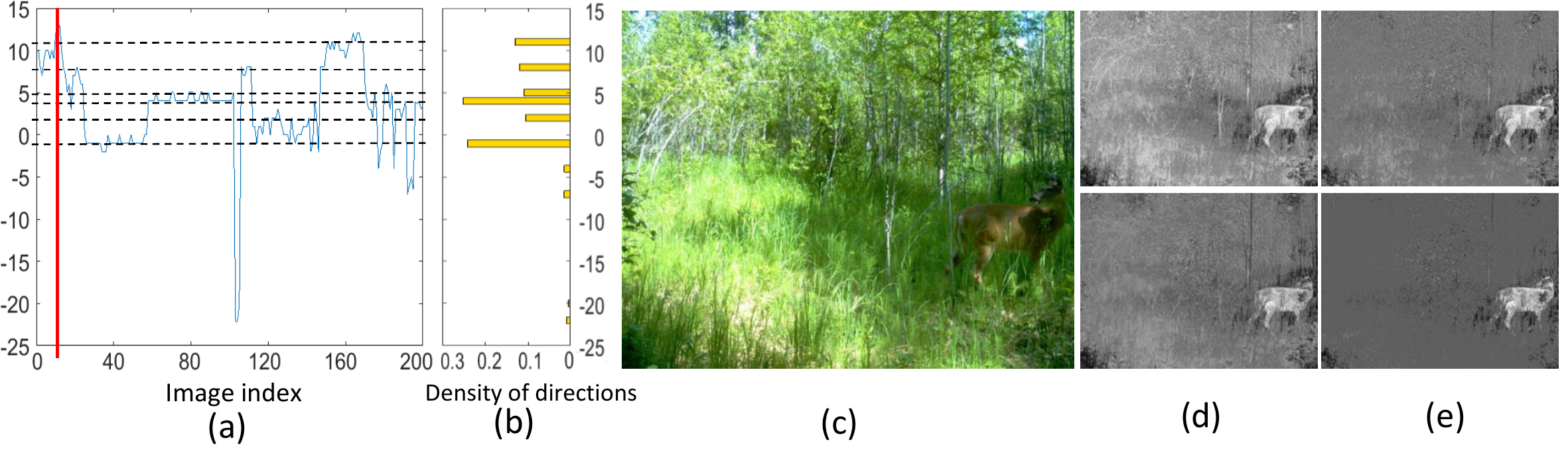}
\vspace{-20pt}
\caption{(a) Best invariant direction of each image in a sequence
 ($y$-axis: angle of the invariant directions $e^{\perp}$ in degrees), (b) Dominant directions (yellow bars) after clustering, (c) $11^{th}$ image in the sequence as shown with a red line in (a), where its best invariant direction is $13^{\circ}$, (d) The first and the second rows show the invariant representations of the selected image using the average direction of the sequence ($5^{\circ}$) and its best direction ($13^{\circ}$), respectively. (e) Obtained outliers of the invariant representations.}
\label{invariant_sample}
\vspace{-7pt}
\end{figure}
Fig.~\ref{invariant_sample} shows an example of the variability of the illumination invariant direction in an image sequence and its impact on generating a illumination-invariant image representation. Fig.~\ref{invariant_sample}(a) shows the invariant directions of an image sequence of 200 frames while illumination changes (blue line), one direction for each image, varying mostly between $-4^o$ and $13^o$. Fig.~\ref{invariant_sample}(c) shows a selected image from the sequence, which is image $11$ and corresponds to the red line in Fig.~\ref{invariant_sample}(a). The invariant direction for this image is found to be $13^{\circ}$ while the average invariant direction of the sequence is around $5^{\circ}$, when we assume $\chi'$ moves exactly along a straight line. Fig.~\ref{invariant_sample}(d) compares the two invariant representations created with invariant directions of $5^{\circ}$ and $13^{\circ}$, respectively, and Fig.~\ref{invariant_sample}(e) shows the detected foreground objects using these two different representations from the RPCA method where the use of the optimal invariant direction ($13^o$) produces much more desirable result than that of the sub-optimal direction ($5^o$). 
This example clearly shows the importance of the choice of the invariant direction in creating the invariant representations, and the undesirable outcome when these representations are created with a sub-optimal invariant direction. 

Our idea to account for the difference in the invariant direction among the images in the sequence, is to first estimate the image-specific invariant directions for the sequence, and then use a clustering algorithm to identify the dominant directions (dotted lines in Fig.~\ref{invariant_sample}(a) or the dominant yellow bars in Fig.~\ref{invariant_sample}(b)). Subsequently, for each image, we create multiple invariant representations, one for each dominant direction, and these multiple representations serve as multiple prior maps for the image. In particular, for each image, we first
use the method in~\cite{IIR_SH} to determine its best invariant direction. With $n_2$ images in an image sequence, this results
in $n_2$ invariant directions where $n_2= 200$ in
Fig.~\ref{invariant_sample}(a). Second, we use k-means to identify $k=10$ clusters
of the $n_2$ invariant directions. Third, we choose the centroid
of a cluster as a dominant invariant direction if the cluster
has support by at least $10\%$ of the images (yellow bars in Fig.~\ref{invariant_sample}(b)). By definition,
there are no more than 10 dominant directions.

\begin{figure}[b!]
\vspace{-10pt}
\centering
\includegraphics[width=\linewidth]{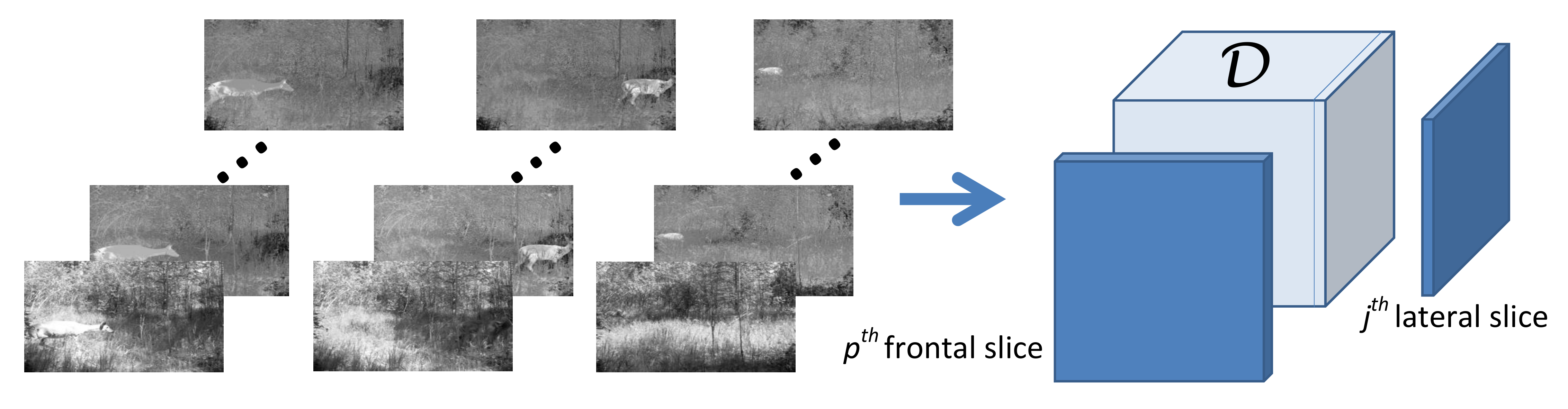}
\caption{Right: sample images with their corresponding illumination invariant representations as prior maps. Left: Tensor $\mathcal{D}$. Frontal slices show $p^{th}$ representation of the images in the sequence. Lateral slices show different representation of each image in the sequence.}
\label{data_structure}
\vspace{-10pt}
\end{figure}
Now, to construct the tensor $\mathcal{D} \in R^{n_1 \times n_2 \times n_3}$ formally (see Fig.~\ref{data_structure}), let ${\mathcal{D}(:,:,1)}$ be an observed image sequence in our problem, where each column of $\mathcal{D}(:,:,1)$ is a vectorized image from the sequence with $n_1$ pixels, and $n_2$ is the number of images in the sequence. $p^{th}$ frontal slice $\mathcal{D}(:,:,p), p= 2,...,n_3$ is a corresponding prior map, generated with a dominant invariant direction. 
Based on this tensor data structure, we are ready to present our new tensor low-rank and invariant sparse decomposition (TLISD) to extract the invariant sparse outliers as moving objects.
\subsection{TLISD Formulation}
\label{sec3_proposed}

\begin{figure}[t]
\centering
\includegraphics[width=\linewidth, height=1.3cm]{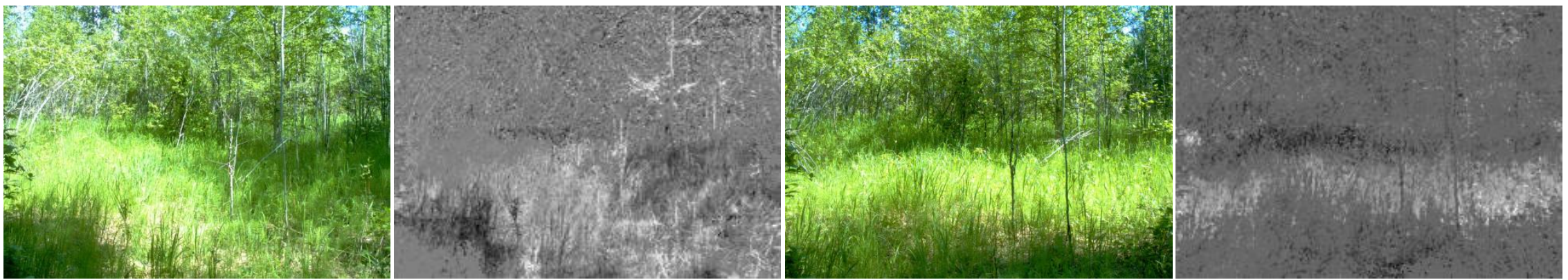}
\vspace{-18pt}\caption{Two sample images and their corresponding illumination changes captured by our proposed method}
\label{sample_illum}
\vspace{-10pt}
\end{figure}

\setlength\emergencystretch{.1\textwidth}
As mentioned in Section~\ref{related}, to detect moving objects under discontinuous illumination change in a sequence, current low-rank methods are insufficient when changes due to illumination and moving shadows are easily lumped with moving objects as the sparse outliers in the low-rank formulation. To separate real changes due to moving objects from those due to illumination, we use multiple prior illumination-invariant maps, introduced in Section~\ref{initial_map}, as constraints on real changes and illumination changes.
In particular, real changes should appear in all frontal slices. Furthermore, lateral slices are completely independent from each other in a time-lapse sequence, but the different representations in each lateral slice (see Fig.~\ref{data_structure}) are from one image and therefore, the locations of real changes should be exactly the same in each lateral slice. 
Now, based on these observations, real changes in each frame should satisfy the group sparsity constraint, which is modeled with the minimization of the $l_{1,1,2}-$norm defined as:
\vspace{-5pt}
\begin{equation}
\vspace{-2pt}
\sum_{i=1}^{n_1} \sum_{j=1}^{n_2} \|\mathcal{S}_{i,j,:} \|_2
\label{eq:group_sparse}
\end{equation}
As discussed, illumination changes in an image sequence should be accounted for by both the low-rank part and the sparse part. The highly correlated part of illumination can be modeled with the low-rank tensor $\mathcal{L}$ as background, but the independent changes in illumination are grouped as the foreground. To capture these uncorrelated illumination and shadow changes, and separate them from real changes, we recognize that they are not necessarily sparse. Fig.~\ref{sample_illum} shows two samples extracted illumination changes using our proposed method. Based on Fig.~\ref{sample_illum}, it is easy to understand that illumination changes are on entire image and so, those uncorrelated changes are not completely sparse. 
These properties can be conveniently modeled with the $k-$support norm~\cite{support_stable}, which is a balanced norm and defined as:
\vspace{-5pt}
\begin{gather}
\vspace{-2pt}
\|\mathcal{C}_{:,:,p}\|_k^{sp}\hspace{-2pt} =\hspace{-2pt} \Big ( \sum_{m=1}^{k-r-1}\hspace{-2pt}(|c|_{m}^{\downarrow})^2+\frac{1}{r+1}(\sum_{m=k-r}^{d}\hspace{-5pt}|c|_m^{\downarrow})^2 \Big )^{\frac{1}{2}}
\label{eq:support_norm}
\end{gather}
where $\mathcal{C}_{:,:,p}$ and $|c|_m^{\downarrow}$ denote the $p^{th}$ frontal slice of $C$ and the $m^{th}$ largest element in $|c|$, respectively. $r \in \{0; 1;...; k-1\}$ is an integer that is computed automatically by Algorithm 2 in the supplementary material. $c=vec({\mathcal{C}_{:,:,p}})$ represents the vector constructed by concatenating the columns of $\mathcal{C}_{:,:,p}$ and $d=n1 \times n2$ is the dimension of the frontal slice. The $k$-support norm has two terms: $l_2$-norm penalty for the large component, and $l_1$-norm penalty for the small components. $k$ is a parameter of the cardinality to achieve a balance between the $l_2$-norm and the $l_1$-norm ($k=n_1$ in our experiments). The $k$-support norm provides an appropriate trade-off between model sparsity and algorithmic stability~\cite{support_stable}, and yields more stable solutions than the $l_1$-norm~\cite{efficient_support}.
In this paper we show that the $k$-support norm can estimate the illumination changes in an image sequence accurately. Joining of this norm and~(\ref{eq:group_sparse}) as two constraints in one optimization framework enables us to separate real changes from illumination changes.
 

To summarize, we propose the tensor low-rank and invariant sparse decomposition (TLISD) method, as follows.
\begin{equation}
\min_{\mathcal{L,S,C}}
\,\|\mathcal{L}\|_* + \lambda_1 \|\mathcal{S}\|_{1,1,2} + \lambda_2 (\|\mathcal{C}\|_k^{sp})^2 \notag
\vspace{-8pt}
\end{equation}
\begin{equation}
s.t. \,\,\,\,\mathcal{D = L + S+ C}
\label{eq:main}
\end{equation}
where $\|\mathcal{L}\|_*$ is the tensor nuclear norm, i.e. the average of the nuclear norm of all the frontal slices ($\|\mathcal{L}\|_*=\frac{1}{n_3}\sum_{p=1}^{n_3}\|\mathcal{L}_{:,:,p}\|_*$), and it approximates the rank of $\mathcal{L}$. $\mathcal{S}$ and $\mathcal{C}$ are detected moving objects and illumination changes, respectively.

\subsection{Optimization Algorithm}
\label{sec3_optimization}

In order to solve~(\ref{eq:main}), we use  the standard inexact augmented Lagrangian method (ALM) with the augmented Lagrangian function $\mathcal{H}(\mathcal{L},\mathcal{S},\mathcal{C},\mathcal{Y};\mu)$  whose main steps are described in this section for completeness.   
\begin{equation}
{\mathcal{H}}(\mathcal{L},\mathcal{S},\mathcal{C},\mathcal{Y};\mu)
 =\|\mathcal{L}\|_*+\lambda_1\|\mathcal{S}\|_{1,1,2}+\lambda_2 (\|\mathcal{C}\|_k^{sp})^2 \notag
\vspace{-5pt}
\end{equation}
\begin{equation}
+  <\mathcal{Y},\mathcal{D}-\mathcal{L}-\mathcal{S}-\mathcal{C}>  
+\frac{\mu}{2}\|\mathcal{D}-\mathcal{L}-\mathcal{S}-\mathcal{C}\|_F^2
\label{eq:lagr}
\end{equation}
where $\mathcal{Y}$ is a Lagrangian multiplier, $\mu$ is a positive auto-adjusted scalar, and $<A,B>=trace(A^TB)$. $\lambda_1=1/\sqrt{max(n_1,n_2)n_3}$ and $\lambda_2$ is a positive scalar. 
Now we solve the problem through alternately updating $\mathcal{L},\mathcal{S},$ and $\mathcal{C}$ in each iteration to minimize $\mathcal{H}(\mathcal{L},\mathcal{S},\mathcal{C},\mathcal{Y};\mu)$ with other variables fixed until convergence as follows.

\vspace{-17pt}
\begin{equation}
\vspace{-15pt}
\mathcal{L}^{t+1} \leftarrow \min_{\mathcal{L}}\|\mathcal{L}\|_*+\frac{\mu}{2}\|\mathcal{L}^t-(\mathcal{D}-\mathcal{S}^t-\mathcal{C}^t+\frac{\mathcal{Y}^t}{\mu})\|_F^2
\label{eq:L}
\end{equation} 
\vspace{3pt}
\begin{equation}
\vspace{-1pt}
\mathcal{S}^{t+1}\hspace{-5pt}\leftarrow\hspace{-2pt}\min_{\mathcal{S}} \lambda_1\|\mathcal{S}\|_{1,1,2}+ \frac{\mu}{2}\|\mathcal{S}^{t}\hspace{-3pt}-(\mathcal{D}\hspace{-3pt}-\mathcal{L}^{t+1}\hspace{-3pt}-\mathcal{C}^t\hspace{-3pt}+\frac{\mathcal{Y}^t}{\mu})\|_F^2
\label{eq:S1}
\end{equation}
\begin{equation}
\vspace{-5pt}
\mathcal{C}^{t+1}\hspace{-5pt}\leftarrow\hspace{-3pt}\min_{\mathcal{C}} \lambda_2 (\|\mathcal{C}\|_k^{sp})^2+ \frac{\mu}{2}\|\mathcal{C}^{t}\hspace{-3pt}-(\mathcal{D}-\mathcal{L}^{t+1}-\mathcal{S}^{t+1}+\frac{\mathcal{Y}^t}{\mu})\|_F^2
\label{eq:C1}
\end{equation}
\begin{gather}
\mathcal{Y}^{t+1} = \mathcal{Y}^{t} + \mu (\mathcal{D}-\mathcal{L}^{t+1}-\mathcal{C}^{t+1}-\mathcal{S}^{t+1})
\label{eq:Y1}
\end{gather}
where $\mu = min (\rho \mu, \mu_{max})$.
Both~(\ref{eq:L}) and~(\ref{eq:S1}) have closed form solutions in~\cite{Lu_tensor} and~\cite{zhang_novel_tensor} respectively, and~(\ref{eq:C1}) has an efficient solution in~\cite{efficient_support}. 
The error is computed as $\|\mathcal{D}-\mathcal{L}^t-\mathcal{S}^t-\mathcal{C}^t\|_F / \|\mathcal{D}\|_F$. The loop stops when the error falls below a threshold ($10^{-5}$ in our experiments). Details of the solutions 
can be found in the supplementary material.
\subsection{Time Complexity}
\label{alg_time}
In this work, we use ADMM to update $\mathcal{L}$ and $\mathcal{S}$, which have closed form solutions. In these two steps the main cost per-iteration lies in the update of $\mathcal{L}_{t+1}$, which requires computing FFT and $n_3$ SVDs of $n_1 \times n_2$ matrices. Thus, time complexity of the first two steps per-iteration is $O(n_1n_2n_3\text{log}n_3+n_{(1)}n_{(2)}^2n_3)$, where $n_{(1)}=max(n_1,n_2)$ and $n_{(2)}=min(n_1,n_2)$~\cite{Lu_tensor}. To update $\mathcal{C}_{t+1}$, we use an efficient solution based on binary search where the time complexity is reduced to $O((n_1n_2+k)\text{log}(n_1n_2))$ for each frontal slice per-iteration~\cite{efficient_support}. Therefore, the total time complexity of the optimization problem~(\ref{eq:main}) is $O(n_1n_2n_3\text{log}n_3+n_{(1)}n_{(2)}^2n_3+(n_1n_2+k)n_3\text{log}(n_1n_2))$.
\section{Experimental Results and Discussion}
\label{experiment}
In this section, we provide an experimental evaluation of our proposed method, TLISD. We first evaluate the effect of each term in~(\ref{eq:main}) and their $\lambda$ coefficients. Then, we evaluate TLISD on benchmark frame-rate image sequences or those that are captured via time-lapse or motion-triggered photography. 
We also introduce a new dataset captured by industrial security cameras and wildlife monitoring systems during three years, and evaluate our method on this dataset.
\subsection{Experiment Setup}
\label{sec:exp.set}

\noindent{\it{\textbf{Existing datasets}}}: We evaluate our TLISD method on eleven selected sequences from the CDnet dataset~\cite{exp_cdnet}, Wallflower dataset~\cite{exp_wallflower}, I2R dataset~\cite{exp_i2r}, and ICD~\cite{shakeri_ICCV}, which include illumination change and moving shadows. 

\noindent {\it{\textbf{Extended Illumnation Change (EIC) dataset:}}} Due to the lack of a comprehensive dataset with various illumination and shadow changes in a real environment, we have created a new benchmark dataset called EIC with around 80k images in 15 sequences, captured via available surveillance systems in wildlife and industrial applications. Particularly, ten sequences are captured via wildlife monitoring systems, and five sequences from industrial applications, with three railway sequences and two construction site sequences. Six sample sequences of this dataset are shown in Fig.~\ref{qualitative_ours}. All sequences can be found in the supplementary material.

{
\setlength\emergencystretch{.1\textwidth}
\noindent{\it{\textbf{Evaluation metric}}}: For quantitative evaluation, pixel-level F-measure $=$ $ 2\frac{recall \times precision}{recall + precision}$ is used. We also compare the different methods in execution time in seconds.  
}
\subsection{Algorithm Evaluation: The effect of term $\mathcal{C}$}
\label{alg_eval_terms}
\begin{figure}[b!]
\vspace{-15pt}
    \begin{subfigure}[b]{0.49\linewidth}
        \includegraphics[width=\linewidth, height=2.5cm]{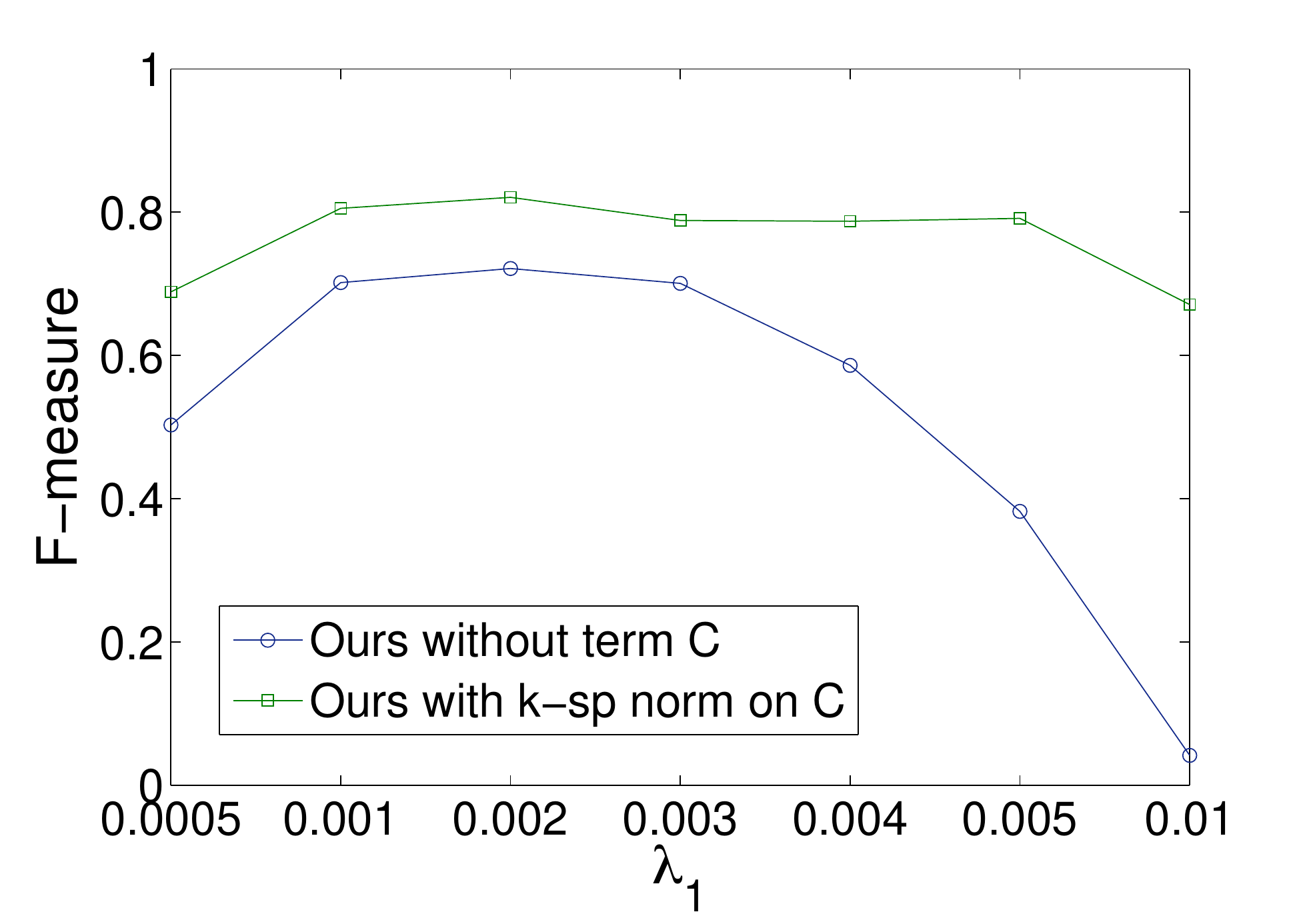}\hspace{-15pt}
        \vspace{-9pt}\caption{}\vspace{-4pt}
    \end{subfigure}
   \begin{subfigure}[b] {0.49\linewidth}\hspace{-5pt}
        \includegraphics[width=\linewidth, height=2.5cm]{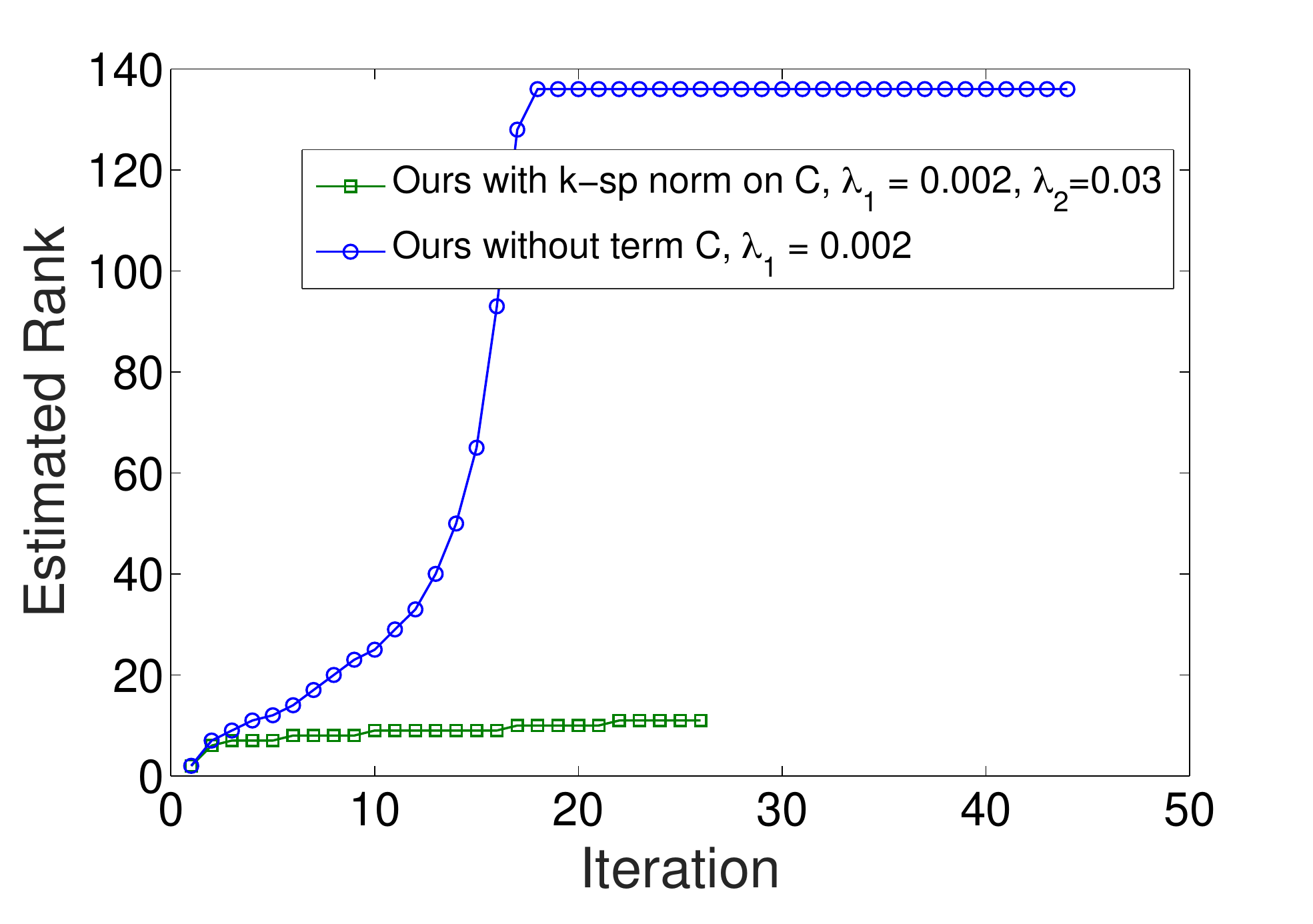}
        \vspace{-9pt}\caption{}\vspace{-4pt}
    \end{subfigure}
    ~
    \begin{subfigure}[b]{0.49\linewidth}
        \includegraphics[width=\linewidth, height=2.5cm]{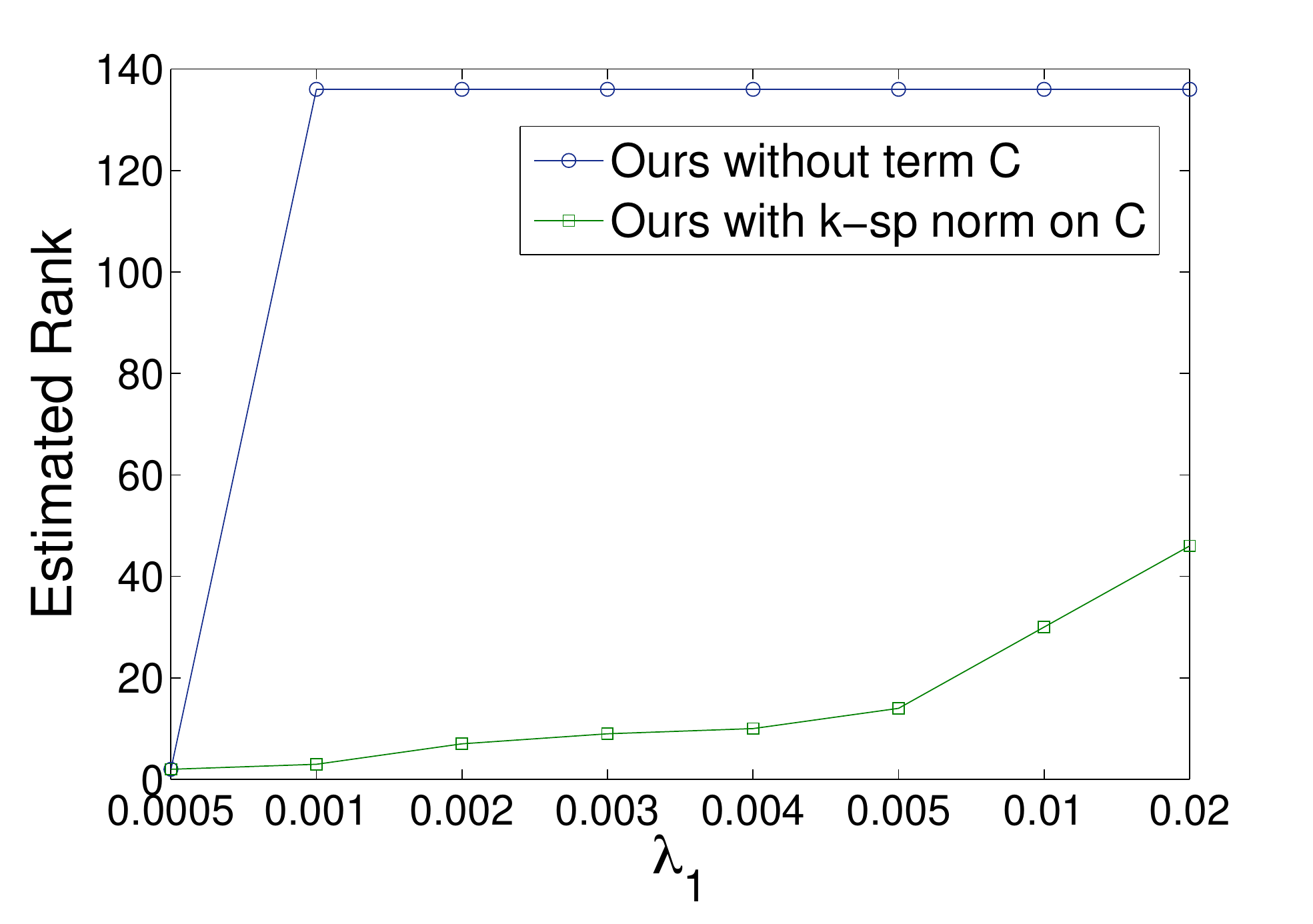}\hspace{-15pt}
        \vspace{-8pt}\caption{}\vspace{-4pt}
    \end{subfigure}
    \begin{subfigure}[b]{0.49\linewidth}\hspace{-5pt}
        \includegraphics[width=\linewidth, height=2.5cm]{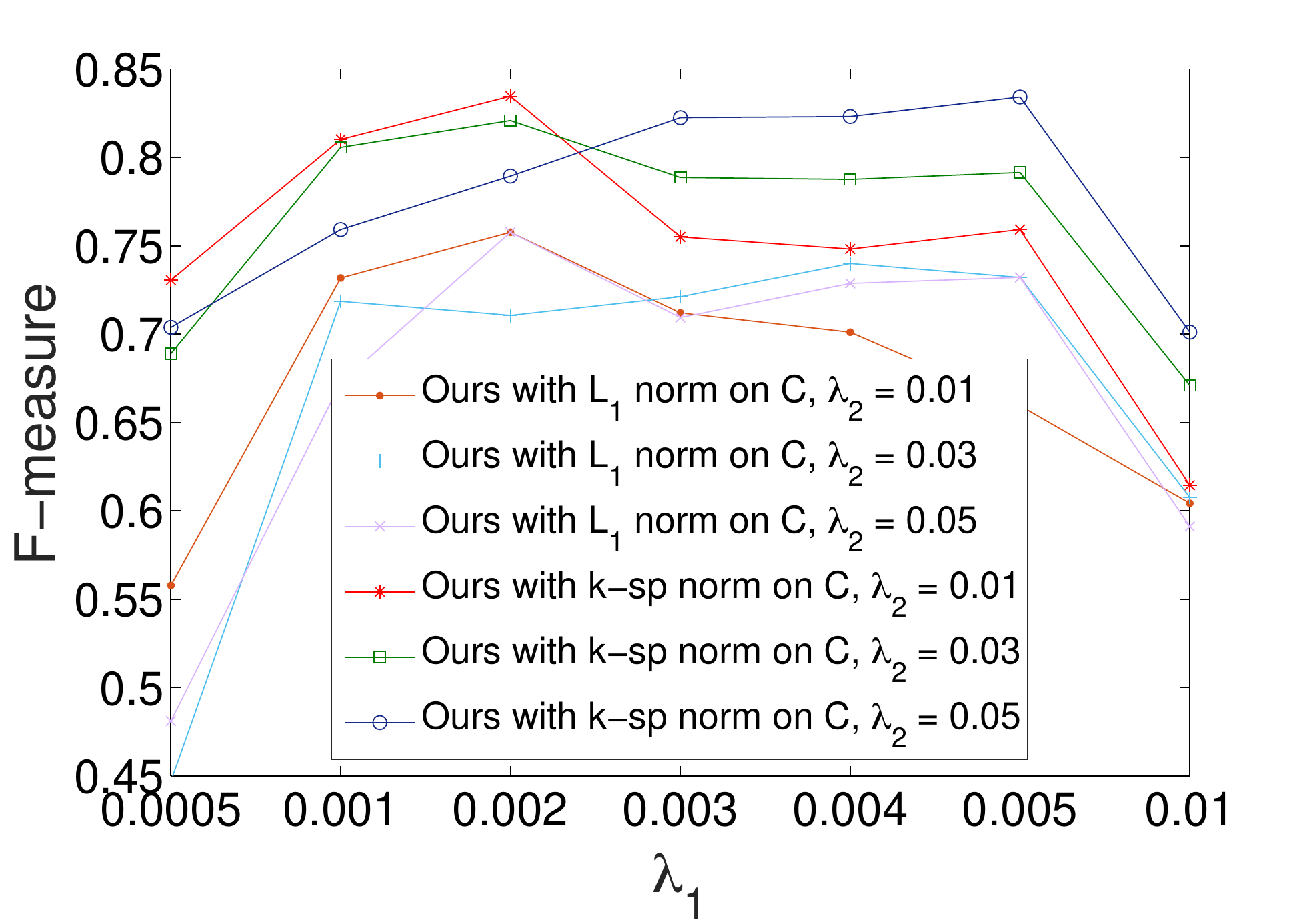}
        \vspace{-8pt}\caption{}\vspace{-4pt}
    \end{subfigure}
    ~
    \begin{subfigure}[b]{0.49\linewidth}
        \includegraphics[width=\linewidth, height=2.5cm]{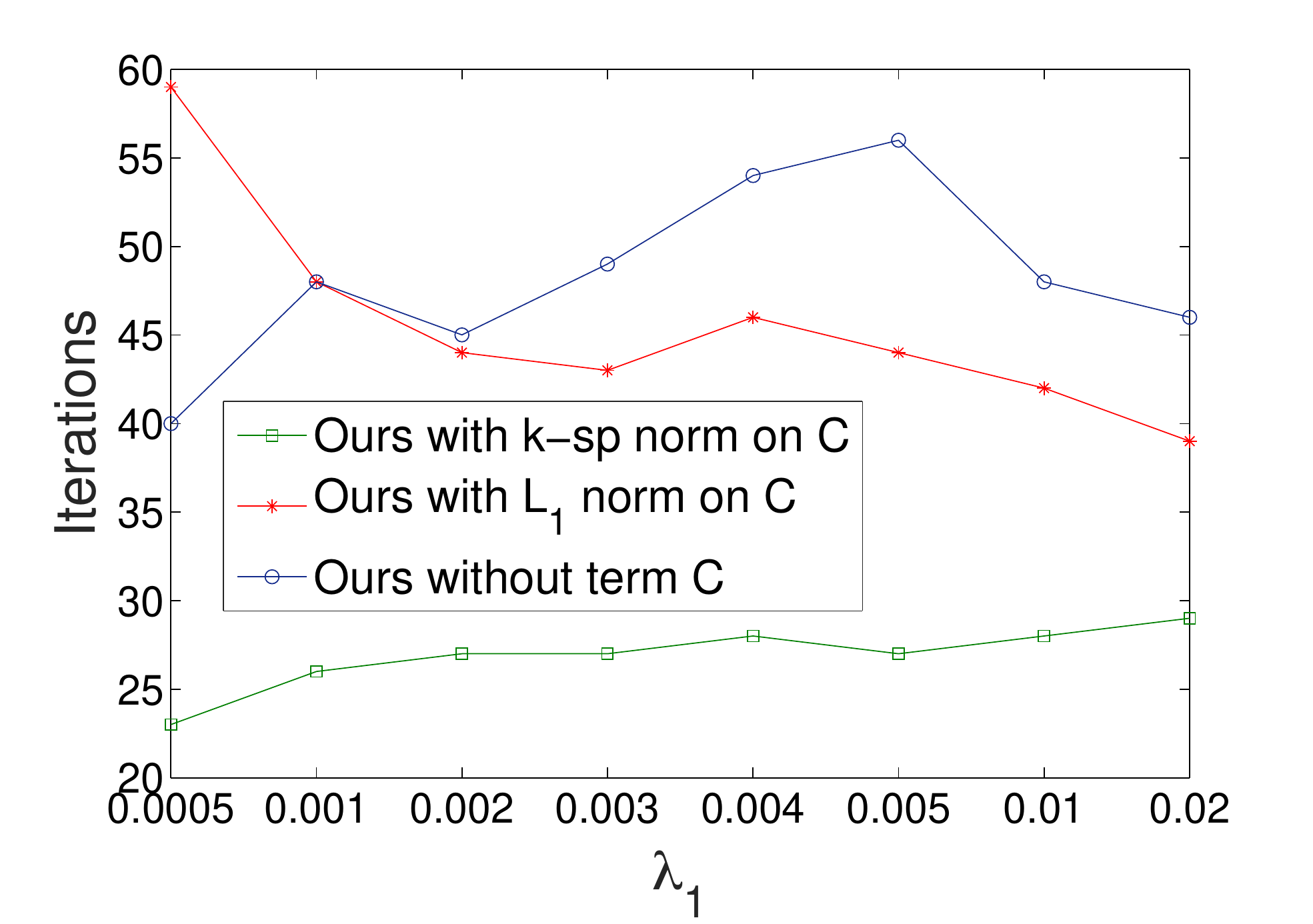}\hspace{-15pt}
        \vspace{-8pt}\caption{}
    \end{subfigure}
    \begin{subfigure}[b]{0.495\linewidth}\hspace{-4pt}
        \includegraphics[width=\linewidth, height=2.5cm]{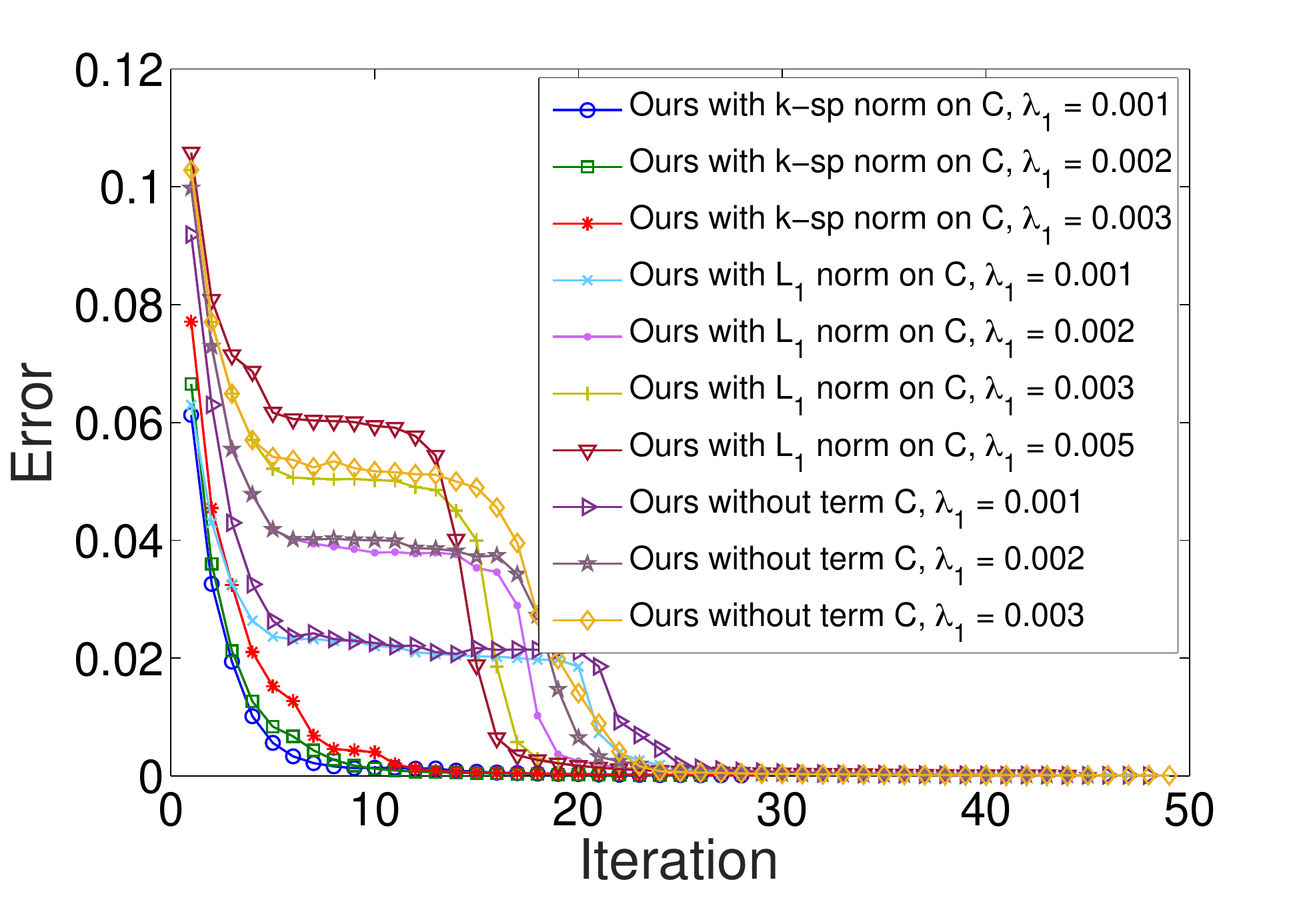}
        \vspace{-8pt}\caption{}
    \end{subfigure}
    \vspace{-20pt}
    \caption{Self evaluation of TLISD. (a) Average F-measure with different values for $\lambda_1$ on all ICD sequences between TLISD and~(\ref{eq:WO_C}), (b) Estimated rank of TLISD and~(\ref{eq:WO_C}) through iterations on sequence ``Wildlife3", (c) Estimated rank of sequence ``Wildlife3" with different values for $\lambda_1$, (d) Average F-measure with different values for $\lambda_1$ and $\lambda_2$ on all ICD sequences between TLISD and~(\ref{eq:W_C}), (e) Average number of iterations to converge TLISD,~(\ref{eq:WO_C}) and~(\ref{eq:W_C}) on all ICD sequences, (f) Convergence curves of minimization error for TLISD,~(\ref{eq:WO_C}) and~(\ref{eq:W_C}) on sequence ``Wildlife3".}
    \label{self_eval}
    \vspace{-5pt}
\end{figure}
In the first set of experiments, we evaluate 
the effect of term $\mathcal{C}$ in TLISD when we set different values for $\lambda_1$, in comparison with TLISD without term $\mathcal{C}$, where~(\ref{eq:main}) becomes
\vspace{-3pt}
\begin{equation}
\min_{\mathcal{L,S}}
\,\|\mathcal{L}\|_* + \lambda_1 \|\mathcal{S}\|_{1,1,2}
\hspace{10pt}s.t. \,\,\,\,\mathcal{D = L + S}
\label{eq:WO_C}
\vspace{-3pt}
\end{equation}


Fig.~\ref{self_eval}(a) shows~(\ref{eq:WO_C}) can achieve around $70\%$ accuracy with a well-tuned $\lambda_1=0.002$. Although the result shows the importance of multiple priors and the effect of group sparsity on them, the accuracy of~(\ref{eq:WO_C}) is still far below the accuracy of proposed TLISD by at least $10\%$, even with a well-tuned $\lambda_1$. Fig.~\ref{self_eval}(a) also shows that adding term $\mathcal{C}$ and $k-support$ norm increases the robustness of our algorithm against tuning $\lambda_1$. In fact, in~(\ref{eq:WO_C}) all illumination variations would be assigned to either of $\mathcal{L}$ or $\mathcal{S}$. In this case, those variations should be assigned to the background ($\mathcal{L}$); however, they do not actually belong to background (e.g. moving shadows). As a result, the rank would be increased to absorb these changes into $\mathcal{L}$ and naturally some parts of the moving objects $\mathcal{S}$ would be also absorbed into the background. Fig.~\ref{self_eval}(b) supports the conclusion and shows the obtained rank through the iterations of the optimization. Between iterations 15 and 20, the rank of our method without term $\mathcal{C}$ significantly increases to absorb all variations into $\mathcal{L}$, and to complete the conclusion, Fig.~\ref{self_eval}(f) shows that around the same iterations, the residual error of the method without term $\mathcal{C}$ is significantly reduced. This means, illumination variations and shadow changes must grouped into either of $\mathcal{L}$ or $\mathcal{S}$, for~(\ref{eq:WO_C}) to converge. Estimated rank in Fig~\ref{self_eval}(c) shows the proof of this concept. Obviously, with a very small $\lambda_1$, the estimated rank of $\mathcal{L}$ for~(\ref{eq:WO_C}) is small and all illumination variations are easily lumped with moving objects in $\mathcal{S}$. This causes less accuracy and sometimes even cannot provide meaningful results. In contrast, TLISD can estimate a balanced rank and classify illumination variations into term $\mathcal{C}$ with $k-support$-norm on it instead of increasing the rank to absorb them into $\mathcal{L}$.

To justify the use of $k-support$ norm on $\mathcal{C}$ in TLISD, we also compare the method with the other potential term on $\mathcal{C}$, which is $l_1$-norm to absorb outliers, i.e., define~(\ref{eq:main}) as
\vspace{-3pt}
\begin{equation}
\min_{\mathcal{L,S,C}}
\,\|\mathcal{L}\|_*\hspace{-2pt} +\hspace{-2pt} \lambda_1 \|\mathcal{S}\|_{1,1,2}\hspace{-2pt} +\hspace{-2pt}  \lambda_2 \|\mathcal{C}\|_{1}
\hspace{2pt}s.t. \,\mathcal{D\hspace{-2pt} = \hspace{-2pt}L\hspace{-2pt} +\hspace{-2pt} S +\hspace{-2pt} C}
\label{eq:W_C}
\vspace{-3pt}
\end{equation}

For this experiment, we evaluate our method with both $l_1$ and $k-support$ norms on $\mathcal{C}$ under different values of $\lambda_1$ and $\lambda_2$. Fig.~\ref{self_eval}(d) illustrates the accuracy of our method with either of regularizers. Although $l_1$-norm can increase the accuracy and robustness of the moving object detection in comparison with~(\ref{eq:WO_C}) that we showed in Fig.~\ref{self_eval}(a), the obtained accuracy is still less than TLISD. In addition, the number of iterations to converge, for both~(\ref{eq:WO_C}) and~(\ref{eq:W_C}) is much more than that of in TLISD.  Fig.~\ref{self_eval}(e) shows the average number of iterations for all three possible methods with different setup for $\lambda_1$ on all ICD sequences. For both TLISD and~(\ref{eq:W_C}), $\lambda_2 = 0.03$, which produces robust results over different values of $\lambda_1$(refer to Fig.~\ref{self_eval}(d)). As discussed in Section~\ref{proposed}, illumination changes are not necessarily sparse and can be found throughout an image. Therefore, $l_1$-norm is not a suitable regularizer to capture illumination changes. In such cases, the same issue as~(\ref{eq:WO_C}) happens when the optimizer increases the rank to minimize the residual error. Fig.~\ref{self_eval}(f) shows the error of all three methods through iterations. For~(\ref{eq:W_C}), the same pattern as~(\ref{eq:WO_C}) is seen to decrease the error while the rank increases through optimization.\vspace{-1pt}

\subsection{Evaluation on Benchmark Sequences}
\begin{figure}[b]
\vspace{-5pt}
\centering
\includegraphics[width=\linewidth]{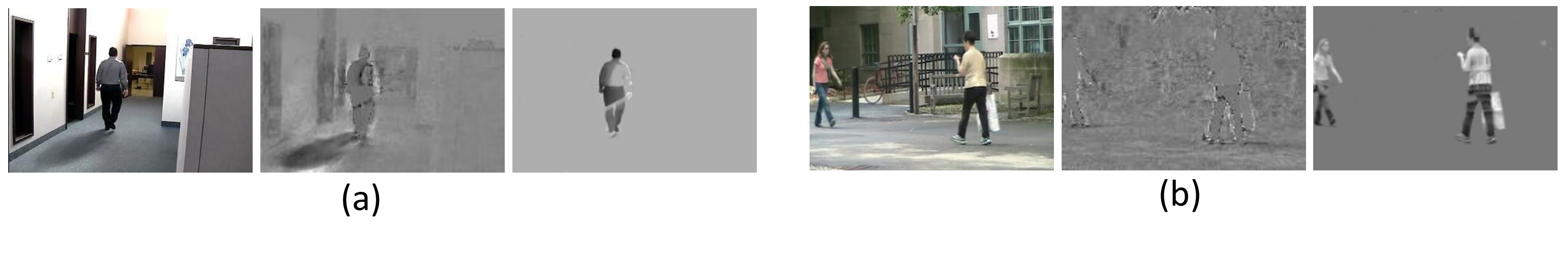}
\vspace{-25pt}\caption{Columns from left to right show sample image, illumination changes, and detected moving objects for (a) cubicle and (b) backdoor sequences}
\label{res_benchmark_sample}
\end{figure}

In this section we evaluate our method on the eleven benchmark sequences described in Section~\ref{sec:exp.set}.~Fig.~\ref{res_benchmark_sample} shows the qualitative results of TLISD on ``Cubile" and ``Backdoor". The second and the third columns of Figs.~\ref{res_benchmark_sample}(a) and (b) illustrate the first frontal slice of $\mathcal{C}$ and $\mathcal{S}$, corresponding to illumination changes and moving objects, respectively. The high-quality of our detection result $\mathcal{S}$ is clearly visible. 

\begin{figure}[t]
\centering
\includegraphics[width=\linewidth, height=4cm]{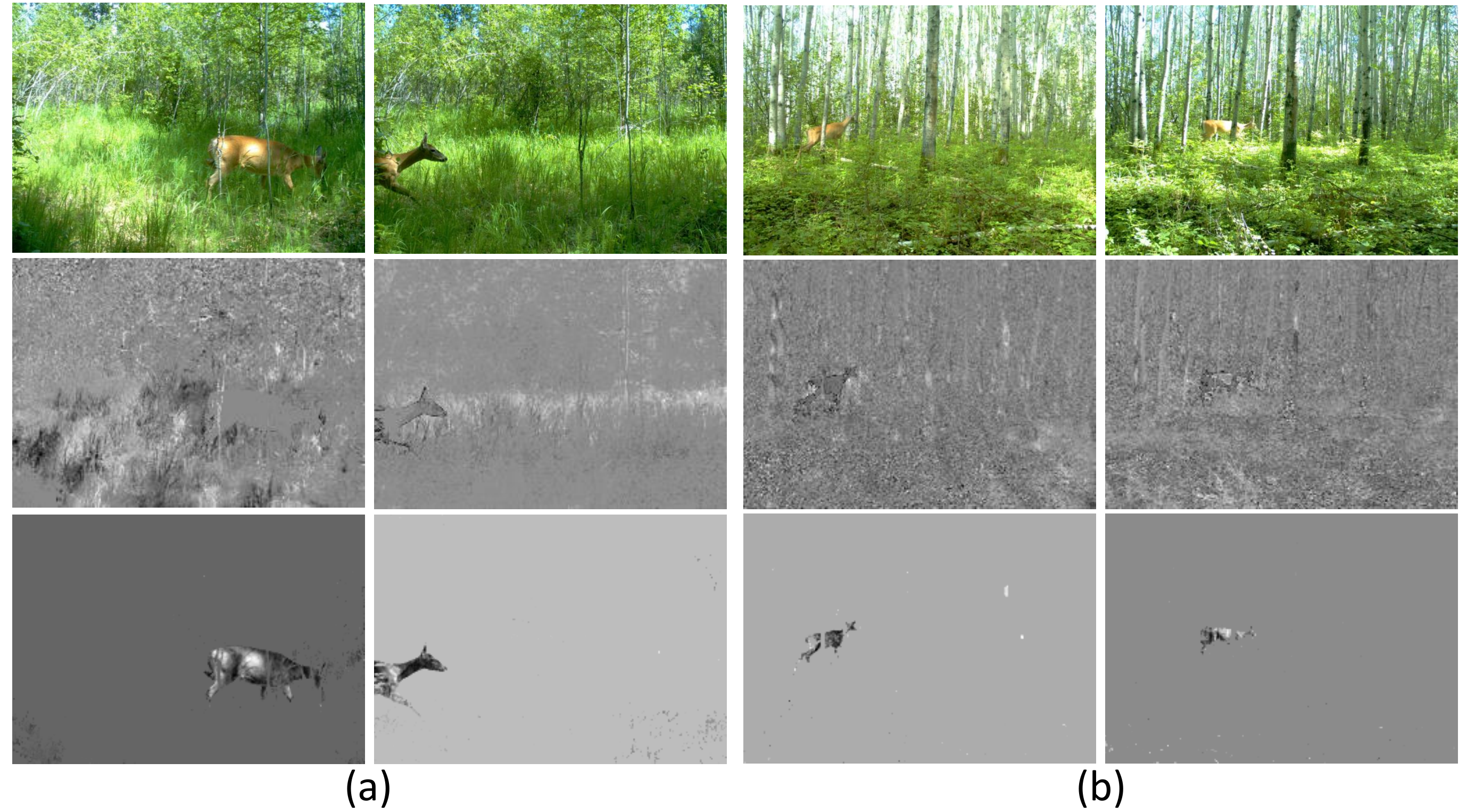}
\vspace{-20pt}
\caption{First row: two sample images from (a) Wildlife1, (b) Wildlife3 sequences. Second row: illumination changes obtained from the first frontal slice of $\mathcal{C}$. Third row: detected objects from the first frontal slice $\mathcal{S}$.}
\label{qualitative_1}
\vspace{-10pt}
\end{figure}

Figs.~\ref{qualitative_1}(a) and (b) show qualitative results of our method on two sample sequences of ICD, which has the most challenging conditions in terms of illumination changes. To appreciate the significant variations of illumination we show two images from each sequence. The second and the third rows of each sub-figure show the first frontal slice of $\mathcal{C}$ and $\mathcal{S}$, respectively. The results show the proposed method can accurately separate the changes caused by illumination and shadows from real changes.

We then compare TLISD quantitatively with two online and eight related RPCA batch methods. From online methods we select GMM~\cite{GMM_new} as a baseline method and GRASTA~\cite{GRASTA_new} as an online method that uses the framework of low-rank and sparse decomposition. Also among batch methods, we select SSGoDec~\cite{int_ssgodec}, PRMF~\cite{int_prmf}, PCP~\cite{int_low_02}, Markov BRMF~\cite{int_brmf}, DECOLOR~\cite{int_low_08}, LSD~\cite{int_low_10}, ILISD~\cite{shakeri_ICCV}, and TRPCA~\cite{Lu_tensor}. For all the competing methods we use their original settings through LRS Library~\cite{lrslibrary}, which resulted in the best performance. For quantitative evaluation of RPCA-related methods, a threshold criterion is required to get the binary foreground mask. Similarly, we adopt the same threshold strategy as in~\cite{lrslibrary}. In TLISD, $\lambda_1=1/\sqrt{max(n_1,n_2)n_3}$ (similar to TRPCA) and $\lambda_2=0.03$. 
Table~\ref{table_benchmark} shows the performance of TLISD in comparison with the competing methods in terms of F-measure. For all the sequences TLISD ranked among the top two of all methods, and achieves the best average F-measure in comparison with all other methods. Although DECOLOR, LSD, and ILISD work relatively well, Only ILISD is comparable with our method due to the use of illumination regularization terms in ILISD. This evaluation shows the effectiveness of multiple prior maps and $k-support$ norm as two regularization terms for separating moving objects from illumination changes, and boosting the overall performance of object detection.

\setlength{\tabcolsep}{4pt}
\begin{table*}[t]
\begin{center}
 \resizebox{0.9\linewidth}{!}{
  \begin{tabular}{|l|c|c|c|c|c|c|c|c|c|c|c|}
  \hline
    Sequence     & Backdoor & CopyMachine   & Cubicle  & PeopleInShade   & LightSwitch    & Lobby & Wildlife1        & Wildlife2          & Wildlife3           & WinterStreet           & MovingSunlight\\
    \hline
    GMM~\cite{GMM_new} & 0.6512 & 0.5298 & 0.3410 & 0.3305 & 0.4946 & 0.3441 & 0.2374 & 0.2880 & 0.0635 & 0.1183 & 0.0717 \\
    GRASTA~\cite{GRASTA_new} & 0.6822 & 0.6490 & 0.4113 & 0.5288 & 0.5631 & 0.6727 & 0.3147 & 0.3814 & 0.2235  & 0.2276  & 0.1714 \\
    SSGoDec~\cite{int_ssgodec}     & 0.6611  & 0.5401 &  0.3035  & 0.2258 & 0.3804 & 0.0831 & 0.2912  & 0.2430  & 0.0951 & 0.1215 & 0.2824 \\
    PRMF~\cite{int_prmf}  & 0.7251  & 0.6834 &  0.3397  & 0.5163 & 0.2922 &  0.6256 & 0.2718 & 0.3991 & 0.07012 & 0.2108 & 0.2932 \\
    DECOLOR~\cite{int_low_08}      & 0.7656  & 0.7511 &  0.5503  & 0.5559 & 0.5782 & \underline{0.7983} & 0.3401 & 0.3634 & 0.1202 & 0.4490 & 0.3699 \\
    PCP~\cite{int_low_02}& 0.7594  & 0.6798 &  0.4978  & 0.6583 & \textbf{0.8375} & 0.6240 & 0.5855 & 0.6542 &  0.3003 & 0.1938 & 0.3445 \\
    BRMF~\cite{int_brmf}   & 0.6291  & 0.3293 &  0.3746  & 0.3313 & 0.2872 & 0.3161 & 0.2743 & 0.2812 & 0.0735 & 0.0872 & 0.2408 \\
    LSD~\cite{int_low_10}  & 0.7603  & 0.8174 &  0.4233  & 0.6168 & 0.6640 & 0.7313 & 0.6471 & 0.3790 & 0.0871 & 0.1604 & 0.3593 \\
    ILISD~\cite{shakeri_ICCV}        & \underline{0.8150}  & \underline{0.8179} &  \underline{0.6887}  & \textbf{0.8010} & 0.7128 & 0.7849 & \underline{0.8033} & \underline{0.7277} & \underline{0.7398} & \underline{0.6931} & \underline{0.6475} \\
    TRPCA~\cite{Lu_tensor}  & 0.7022 & 0.6805 & 0.5329 & 0.5683 & 0.6924 & 0.6176 & 0.4382 & 0.3926 & 0.2854 & 0.2721 & 0.3018 \\
    TLISD & \textbf{0.8276} & \textbf{0.8445} & \textbf{0.7350} & \underline{0.7961} & \underline{0.7429} & \textbf{0.8012} & \textbf{0.8862} & \textbf{0.8065} & \textbf{0.8010} & \textbf{0.7092} & \textbf{0.7122} \\
    \hline
  \end{tabular}}
\end{center}
\vspace{-10pt}
\caption{Comparison of F-measure score between our proposed method and other compared methods on benchmark real-time sequences (best F-measure: bold, second best F-measure: underline)}
\label{table_benchmark}
\vspace{-10pt}
\end{table*}
\setlength{\tabcolsep}{1.4pt}

\subsection{Evaluation of TLISD on EIC Dataset}
\label{our_dataset}
\begin{figure}[t!]
\centering
\includegraphics[width=\linewidth, height=8cm]{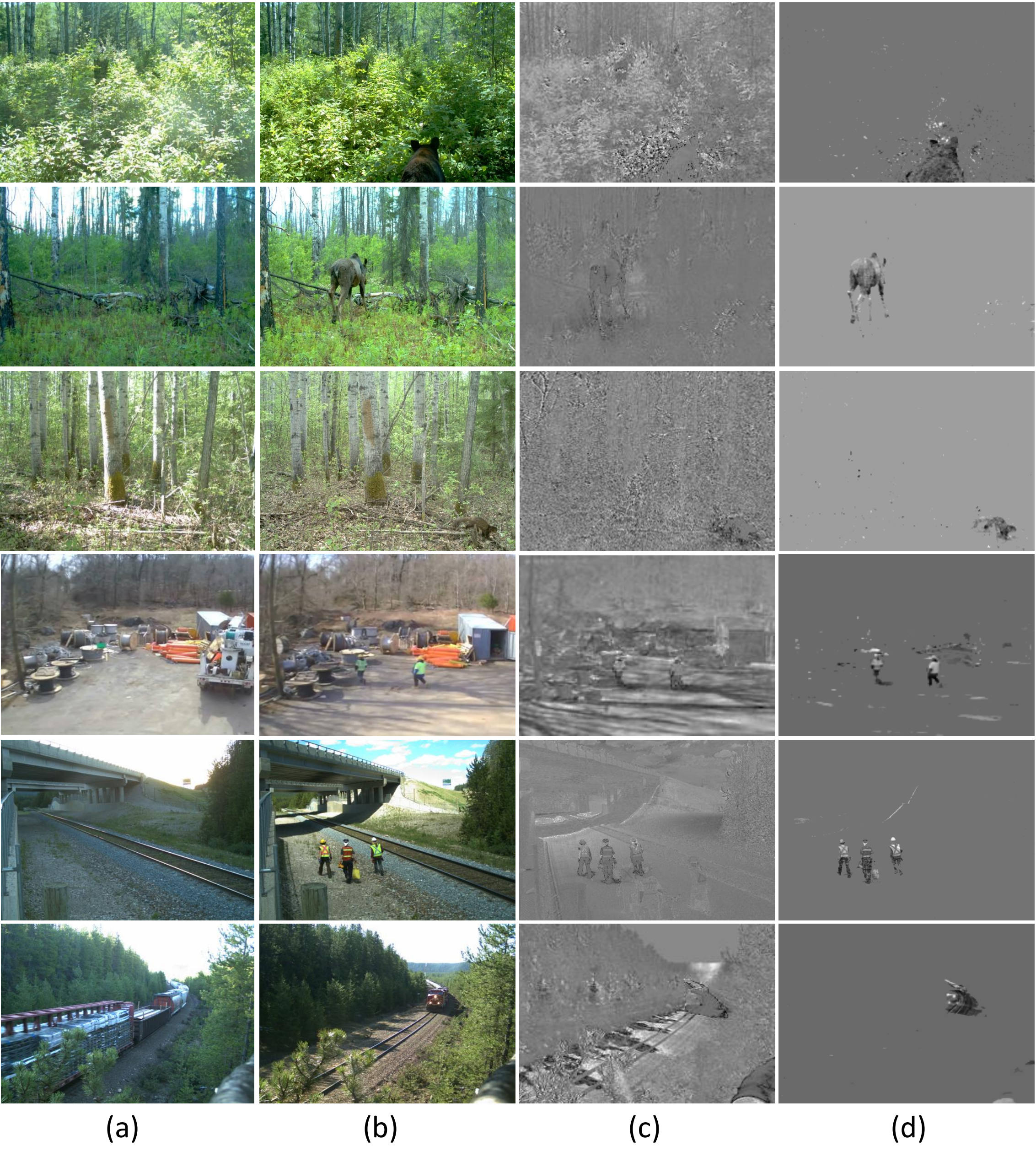}
\vspace{-20pt}
\caption{Columns (a) and (b): two sample images of each sequence, (c) and (d): illumination changes captured in $\mathcal{C}$, and detected objects of images in (b), respectively}
\label{qualitative_ours}
\vspace{-10pt}
\end{figure}
\begin{figure}[t]
\centering
\includegraphics[width=\linewidth, height=4.5cm]{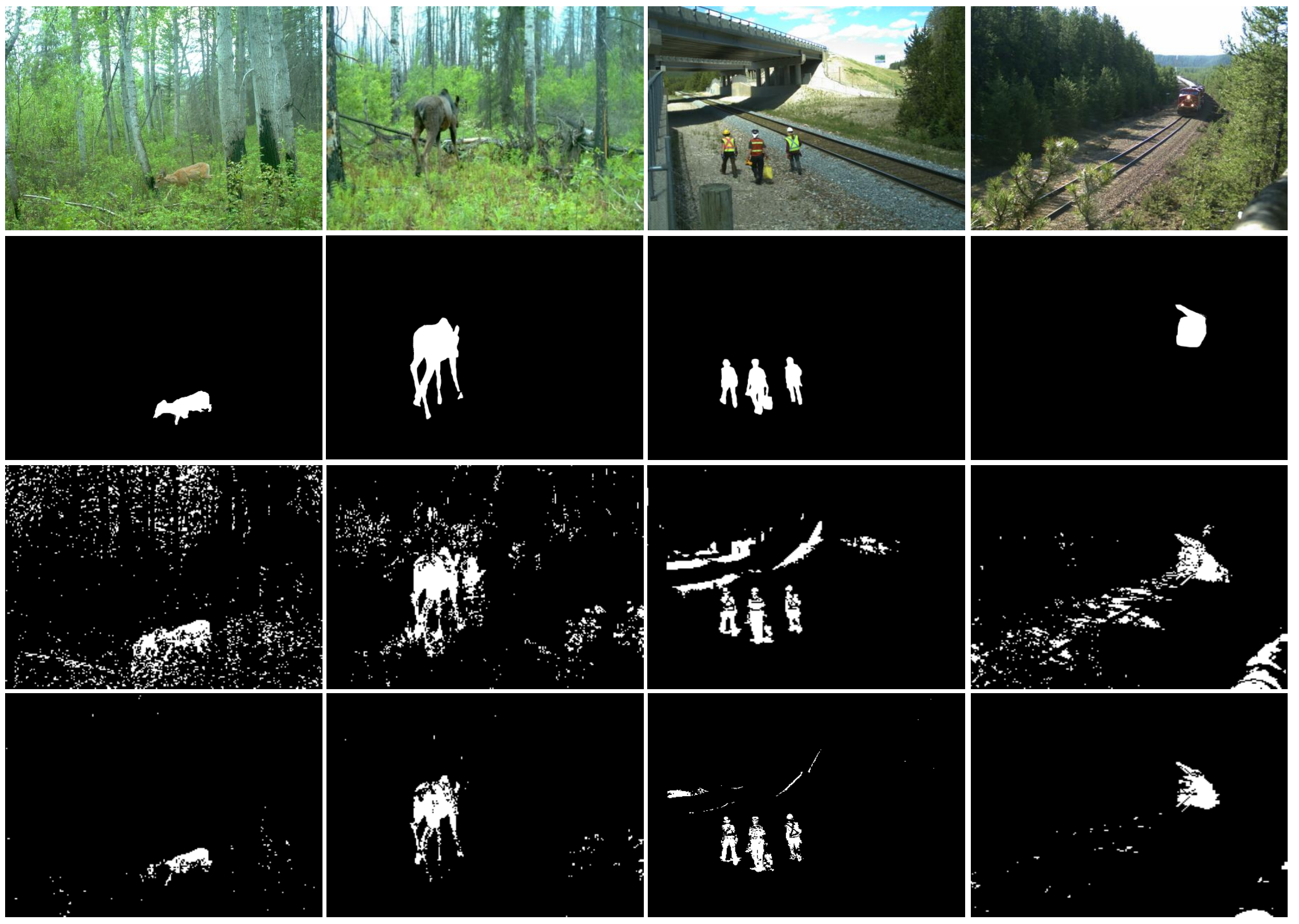}
\vspace{-15pt}
\caption{Comparison of qualitative results between TLISD and ILISD on four sequences of EIC dataset. Top to bottom: Sample Image, Ground Truth, ILISD, and TLISD}
\label{comparison_qualitative_ours}
\vspace{-8pt}
\end{figure}

In this section, we evaluate TLISD on the introduced EIC dataset. Six sample sequences of ELC are shown in Fig.~\ref{qualitative_ours}. To understand the significant variations of illumination and shadow, we show two images from each sequence in Figs.~\ref{qualitative_ours}(a) and (b). Columns (c) and (d) show the first frantal slices of $\mathcal{C}$ and $\mathcal{S}$ obtained by TLISD for the images in column (b), in order to capture illumination changes and to detect moving objects. Table~\ref{table_EIC} show the capability of TLISD in comparison with the four best competitive methods (based on Table~\ref{table_benchmark}) in terms of F-measure, where TLISD can outperform the other methods by a clear performance margin. Fig.~\ref{comparison_qualitative_ours} also compares TLISD with ILISD (the second best method in Table.~\ref{table_EIC}) qualitatively. This qualitative comparison shows that one prior map only is not sufficient for removing the effect of illumination variations and shadow. As discussed in Section~\ref{initial_map}, due to the variation in the invariant direction for images in a sequence, in some conditions separating illumination changes and shadows from real changes is roughly impossible and selecting multiple prior maps is essential. More results on all sequences can be found in the supplementary material. 

\setlength{\tabcolsep}{4pt}
\begin{table}[t]
\begin{center}
 \resizebox{\linewidth}{!}{
  \begin{tabular}{|l|c|c|c|c|c|c|}
  \hline
    Sequence     & Wildlife4 & Wildlife5 & Wildlife6 & Railway1 & Railway2 & Industrial area1\\
    \hline
    PCP~\cite{int_low_02}    & 0.4150 & 0.4016 & 0.3092 & 0.3634 & 0.4086 & 0.2869 \\
    DECOLOR~\cite{int_low_08}& 0.3475 & 0.2010 & 0.2604 & 0.2853 & 0.3021 & 0.3242 \\
    ILISD~\cite{shakeri_ICCV} & 0.6493 & 0.7012 & 0.6501 & 0.6376 & 0.6221 & 0.6089 \\
    TRPCA~\cite{Lu_tensor} & 0.2934 & 0.3082 & 0.2855 & 0.3447 & 0.2805 & 0.2914 \\
    TLISD & \textbf{0.7508} & \textbf{0.8049} & \textbf{0.7522} & \textbf{0.7241} & \textbf{0.7116} & \textbf{0.7035} \\
    \hline
	\end{tabular}}
\end{center}
\vspace{-14pt}
\caption{Comparison of F-measure score between our proposed method and other compared methods on EIC dataset}
\label{table_EIC}
\vspace{-12pt}
\end{table}
\setlength{\tabcolsep}{1.4pt}

\subsection{Execution Time of TLISD}

\begin{figure}[b]
\centering
\vspace{-6pt}
\includegraphics[width=\linewidth, height=3.2cm]{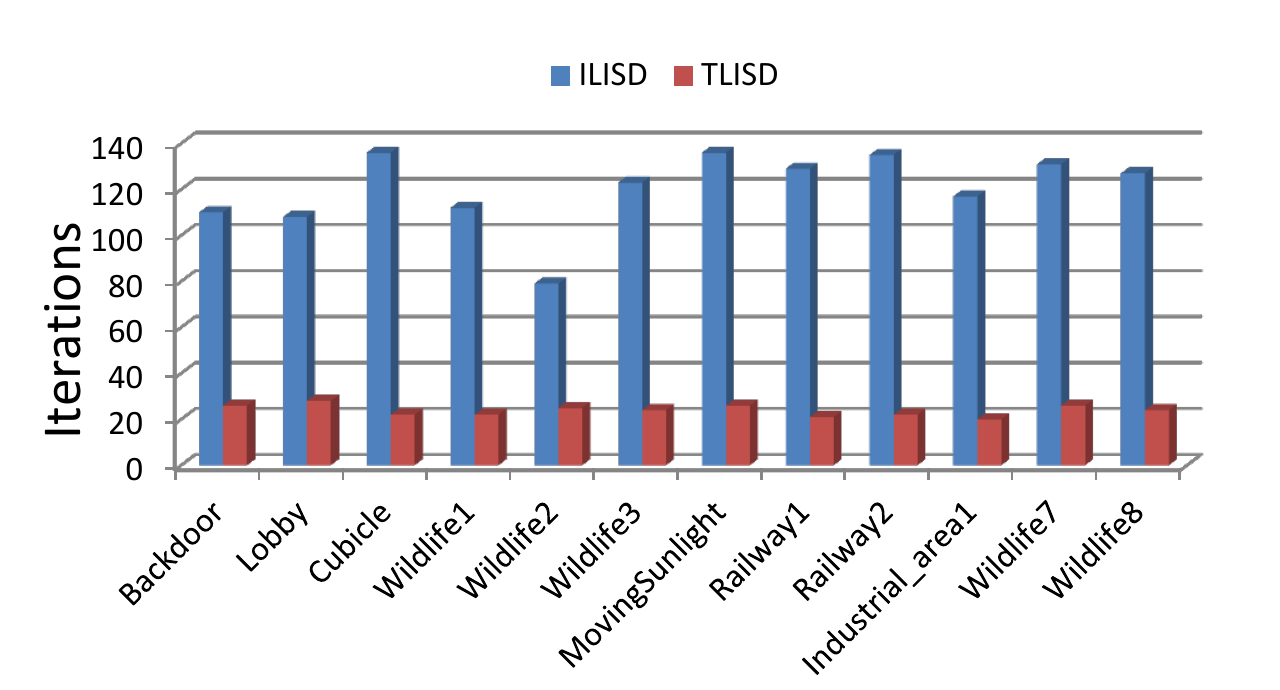}
\vspace{-20pt}
\caption{Number of iterations to converge ILISD and TLISD methods on twelve sequences}
\label{iteration_comparison}
\vspace{-10pt}
\end{figure}

Based on Tables~\ref{table_benchmark} and~\ref{table_EIC}, since ILISD is the only method with comparable results to ours, we examine our proposed method and ILISD in terms of computation time. Table.~\ref{exec_time} compares the execution time of both methods on seven sequences. Regarding the computation time of the proposed method, our tensor-based method needs more time than~\cite{shakeri_ICCV} for each iteration, which is normal due to use of the tensor structure. However, the number of iterations in our method is less than that of ~\cite{shakeri_ICCV}. Fig.~\ref{iteration_comparison} shows the number of iterations to converge for both ILISD and TLISD methods. ILISD~\cite{shakeri_ICCV} has two independent optimization formulae: one for providing a prior map and the other for separating moving objects from illumination changes, and they have independent numbers of iterations to converge. After convergence, the optimized values are interchangeably used in an outer loop, and hence the total number of iterations is much more than that of our method which involves one optimization formula. As discussed in Section~\ref{alg_time}, the dominant time in our method is SVD decomposition for frontal slices, which are independent from each other, and so can be solved in parallel on a GPU to speed up the computation. Therefore, the total time of our method is at least comparable with ILISD and can be even faster due to the fewer number of iterations.

\setlength{\tabcolsep}{4pt}
\begin{table}[b]
\vspace{-10pt}
\begin{center}
 \resizebox{\linewidth}{!}{
  \begin{tabular}{|l|c|c|c|c|c|c|c|}
  \hline
    Sequence     & Backdoor & Lobby & Cubicle & Wildlife1 & Wildlife2  & Wildlife3  & MovingSunlight\\
    \hline
    ILISD        & 0.49     & 0.53  &  0.74   & 1.24      & 1.33       & 1.18       & 2.2           \\
    TLISD        & 0.98     & 2.38  &  1.79   & 2.52      & 4.26       & 4.08       & 5.16          \\
    \hline
  \end{tabular}}
\end{center}
\vspace{-10pt}
\caption{Comparison of execution time (in sec.) per image 
}
\label{exec_time}
\vspace{-10pt}
\end{table}
\setlength{\tabcolsep}{1.4pt}
\section{Conclusions}
\vspace{-2pt}
\label{conclusion}

In this paper, we have proposed a novel method based on tensor low-rank and invariant sparse decomposition to detect moving objects under discontinuous changes in illumination, which frequently happen in video surveillance applications. In our proposed method, first we compute a set of illumination invariant representations for each image as prior maps, which provide us with cues for extracting moving objects. Then we model illumination changes in an image sequence using a $k$-support norm and derive a new formulation to effectively capture illumination changes and separate them from detected foregrounds. Currently, many surveillance systems, especially security and wildlife monitoring cameras, use motion triggered sensors and capture image sequences with significant illumination changes. Our proposed method can solve the problem with a performance that is superior to the state-of-the-art solutions. Our method is also able to extract natural outdoor illumination as labeled data for learning-based methods, which can be an effective alternative to optimization based methods such as ours, but with a sequential formulation, to detect illumination changes and moving objects from image sequences. 
\section*{Acknowledgment}
\vspace{-5pt}
This research is supported in part by NSERC through its Discovery Grant and Strategic Network Grant (NCRN) programs.

{\small
\bibliographystyle{ieee_fullname}
\bibliography{egpaper_final}
}

\onecolumn
\pagebreak

\twocolumn
\begin{table}[h]
\begin{center}
  \textbf{\Large {Supplementary Material}\\
    \large Moving Object Detection under Discontinuous Change in Illumination Using Tensor Low-Rank and Invariant Sparse Decomposition}\\[.2cm]
  Moein Shakeri$^{1}$ and Hong Zhang$^{1}$\\[.1cm]
  {\itshape ${}^1$Department of Computing Science\\ University of Alberta, Edmonton, AB, Canada\\}
  ${}^1$shakeri@ualberta.ca, hzhang@ualberta.ca\\
\end{center}
\end{table}

\setcounter{equation}{0}
\setcounter{figure}{0}
\setcounter{table}{0}
\setcounter{section}{0}
\renewcommand{\theequation}{S\arabic{equation}}
\renewcommand{\thefigure}{S\arabic{figure}}
\renewcommand{\thetable}{S\arabic{table}}
\renewcommand{\thesection}{S\arabic{section}}

In this supplementary document we first provide details of the solutions for our proposed method (TLISD) for equations (8), (9), and (10) in the paper in Section~\ref{solution}. We also provide more experimental results and discussion in Section~\ref{experiment}. Particularly, we show more qualitative results of TLISD and compare them with the other batch competing methods (Table 1 in the paper) in Section~\ref{exp_ICD}. Then we introduce all sequences of our proposed EIC dataset in Section~\ref{exp_EIC}, and show more qualitative and quantitative results of TLISD on them.
\vspace{-2pt}
\makeatletter
\algrenewcommand\ALG@beginalgorithmic{\footnotesize}
\makeatother
\begin{algorithm}[b]
\caption{Tensor Low-rank and Invariant Sparse Decomposition (TLISD)}\label{alg1}
\renewcommand{\Comment}[1]{\State \(\triangleright\) }
\begin{algorithmic}[1]
\Statex {\hspace{-10pt}\textbf{Input:} Tensor data $\mathcal{D}$, Parameters $\lambda_1 = 1/\sqrt{max(n_1,n_2)n_3}$,}
\Statex{$\lambda_2=0.03$, $k = n_1$, $\rho = 1.2$, $\mu = 10^{-3}$}
\While{not converged}
\State $\hspace{-5pt}\mathcal{L}^{t+1}=${\it prox-tnn}$(\mathcal{D}\hspace{-1pt}-\hspace{-1pt}\mathcal{S}^t\hspace{-2pt}-\hspace{-2pt}\mathcal{C}^t+\mu^{-1}\mathcal{Y}^t)$ \label{alg-l1} \hspace{5pt} //solves (8) in the paper
\State {\hspace{-5pt}$\mathcal{S}_{temp}= \mathcal{L}^{t+1}+\mathcal{C}^t-\mathcal{D}+\mu^{-1}\mathcal{Y}^t$}
\State {\hspace{-5pt}for each row $i$ and lateral slice $j$} \label{alg-s1}  \hspace{10pt}//lines 3-5 solve (9) in the paper
\State {\hspace{2pt}$\mathcal{S}^{t+1}(i,j,:)=\Big(1-\frac{\lambda_1}{\mu\|\mathcal{S}_{temp}(i,j,:)\|_F} \Big)_+\mathcal{S}_{temp}(i,j,:)$} \label{alg-s2}
\State {\hspace{-5pt}$\mathcal{C}_{temp} = \mathcal{L}^{t+1}+\mathcal{S}^{t+1}-\mathcal{D}+\mu^{-1}\mathcal{Y}^t$}
\State {\hspace{-5pt}for each frontal slice $p$} \label{alg-c1}  \hspace{30pt} //lines 6-8 solve (10) in the paper
\State {\hspace{2pt}$\mathcal{C}^{t+1}(:,:,p)=k_{}sp(\mathcal{C}_{temp}(:,:,p),k,\mu^{-1}\lambda_2)$} \label{alg-c3}  \hspace{3pt}  //Algorithm 2
\State {\hspace{-5pt}$\mathcal{Y} = \mathcal{Y} + \mu (\mathcal{D} - \mathcal{L}^{t+1} - \mathcal{S}^{t+1} - \mathcal{C}^{t+1})$}
\State {\hspace{-5pt}$\mu = \rho \mu; \hspace{2pt} t = t+1$}
\EndWhile \label{alg1-endwhile}
\Statex \hspace{-10pt}\textbf{Output} {$\mathcal{L}^{t}, \mathcal{S}^{t}, \mathcal{C}^{t}$}
\vspace{10pt}
\Statex {\hspace{-10pt}\textbf{function {\it prox-tnn}($\mathcal{A}$)}}
\State {$\mathcal{M} \leftarrow$ fft($\mathcal{A},[\hspace{2pt}],3$)}
\State {for $i=1 : n_3$}
\State {\hspace{5pt} $[U,S,V] = SVD(\mathcal{M}(:,:,i))$}
\State {\hspace{5pt} $\mathcal{\hat{U}}(:,:,i)=U; \hspace{5pt}\mathcal{\hat{S}}(:,:,i)=S; \hspace{5pt} \mathcal{\hat{V}}(:,:,i)=V$}
\State {\hspace{5pt} Updating {\it t-}$rank$ using soft thresholding operator $\mathcal{\overline{S}}_{(1/\mu)}$}  //Similar to [16]
\State {End for}
\State {$\mathcal{U} \leftarrow$ ifft($\mathcal{\hat{U}}(:,1: ${\it t-}$rank,:),[\hspace{2pt}],3$);} \Statex{$\Sigma \leftarrow$ ifft($\mathcal{\hat{S}}(1:${\it t-}$rank,1:${\it t-}$rank,:),[\hspace{2pt}],3$);}
\Statex{$\mathcal{V} \leftarrow$ ifft($\mathcal{\hat{V}}(:,1:${\it t-}$rank,:),[\hspace{2pt}],3$);}
\State {for $i=1:n_3$}
\State {\hspace{5pt} $\mathcal{X}(:,:,i)=(\mathcal{U}(:,:,i)\Sigma(:,:,i))\mathcal{V}^{T}(:,:,i)$}
\State{End for}
\State {return $\mathcal{X}$}

\end{algorithmic}
\end{algorithm}
\vspace{-5pt}
\section{Details of the solutions for equations (8), (9), and (10)}
\vspace{-5pt}
\label{solution}

All details about TLISD are described in Algorithm~\ref{alg1}. The error is computed as $\|\mathcal{D}-\mathcal{L}^{t}-\mathcal{S}^{t}-\mathcal{C}^{t}\|_F/\|\mathcal{D}\|_F$. The loop stops when the error reaches the value lower than a threshold ($10^{-5}$ in our experiments).  
\makeatletter
\algrenewcommand\ALG@beginalgorithmic{\footnotesize}
\makeatother
\begin{algorithm}[h]
\caption{Solving $k-$support norm}\label{alg2}
\renewcommand{\Comment}[1]{\State \(\triangleright\) }
\begin{algorithmic}[1]
\Statex {\hspace{-10pt}\textbf{function ksp($W,k,\gamma$)}}
\State{$\beta = 1/\gamma$, $\nu=vec(W)$ \hspace{5pt} where $\nu \in R^{d}$, $d=n_1 \times n_2$ //size of each frontal slice}
\State {$z=|\nu|^{\downarrow}, z_0=+\infty, z_{d+1}=-\infty$}
\State {for $r=k-1:0$}
\State {\hspace{5pt} Obtain $l$ by \textbf{BinarySearch(z,k-r,d)}}
\State {\hspace{5pt} $T_{r,l}=\sum_{i=k-r}^l z_i$}
\State {\hspace{5pt} If $\frac{1}{\beta+1}z_{k-r-1} > \frac{T_{r,l}}{l-k+r+1+\beta(r+1)} \geq \frac{1}{\beta+1}z_{k-r}$}
\State {\hspace{15pt} break;}
\State {\hspace{5pt} End If}
\State {End for}
\State {For $i=1:d$}
\State {\hspace{5pt} calculate $q_i$ = \Bigg\{
\begin{tabular}{cc}
$\frac{\beta}{\beta+1}z_i$ & \hspace{18pt}if $i=1,...,k-r-1$ \\
$z_i-\frac{\sum_{i=k-r}^{l}z_i}{l-k+r+1+\beta(r+1)}$ & if $i=k-r,...,l$ \\
$0$  & if $i=l+1,...,d$ 
\end{tabular} 
}
\State {\hspace{5pt} $ w_i = sign(\nu_i)q_i$}
\State {End for}
\State {\textbf{Output : $W$}}
\vspace{10pt}
\Statex {\hspace{-10pt}\textbf{function BinarySearch($z,low,high$)}}
\State {If $z_{low} = 0$}
\State { \hspace{5pt} return $l = low$}
\State {End If}
\State {While $low < high-1$}
\State {\hspace{5pt} $mid=\ulcorner \frac{low+high}{2}\urcorner$} {\hspace{30pt}\scriptsize //$\ulcorner x \urcorner$ represents the smallest integer which is larger than $x$}
\State {\hspace{5pt} If $z_{mid} > \frac{\sum_{i=k-r}^{mid}z_i}{mid-k+r+1+\beta(r+1)}$}
\State {\hspace{15pt} $low=mid$}
\State {\hspace{5pt} Else}
\State {\hspace{15pt} $high=mid-1$}
\State {\hspace{5pt} End If}
\State {End While}
\State {return $l=low$}
\end{algorithmic}
\end{algorithm}

More details and all proofs can be found in [18], [14], and [33]. As mentioned in the paper (Section 4.3), for the fair comparison with TRPCA~[18] we use $\lambda_1 = 1/\sqrt{max(n_1,n_2)n_3}$ for all qualitative and quantitative results of TLISD in Sections 4.3 and 4.4. We also set $\lambda_2=0.03$ based on the obtained results in Section 4.2 (in the paper), where shows the best average result on all sequences. 

\begin{figure*}[t]
\centering
\includegraphics[width=0.9\linewidth, height=16cm]{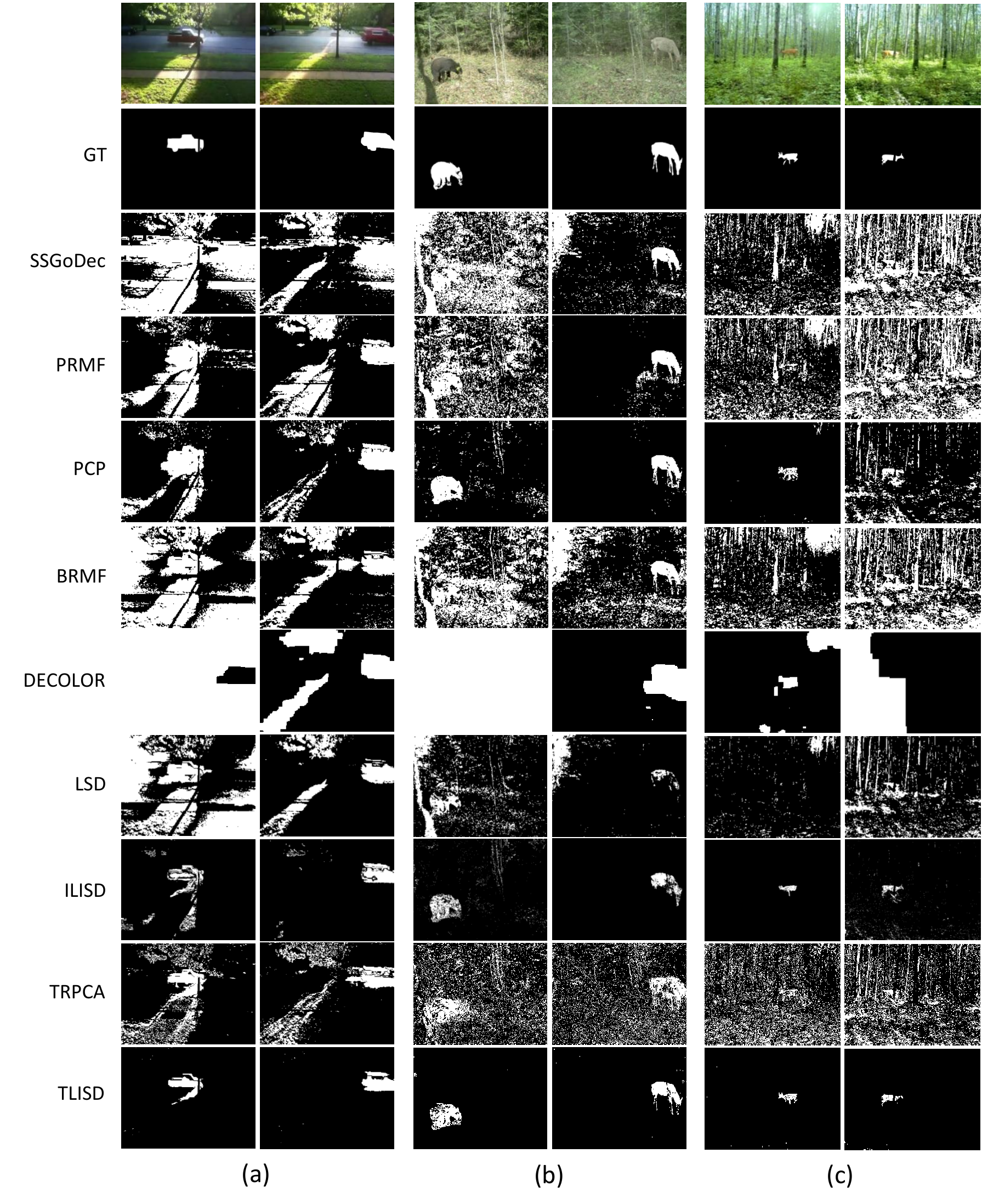}
\caption{Comparison of qualitative results between our method (TLISD) and eight rpca-related methods on two selected images of sequences (a) ``MovingSunLight", (b) ``Wildlife2", and (c) ``Wildlife3"}
\label{fig1}
\end{figure*}

\section{More Experimental Results and Discussion}
\label{experiment}

In this section we show more experimental results of our proposed method and compare them with the results of other competing methods on ICD sequences. We also introduce all sequences of EIC dataset, and show more qualitative and quantitative results on them. 

\subsection{More experimental results on ICD sequences}
\label{exp_ICD}

Fig.~\ref{fig1} shows qualitative comparison of our method with all batch rpca-related methods of Table 1 in the paper. Most methods failed to detect moving objects under significant illumination changes, and only the results of ILISD are comparable with ours. However, based on this qualitative results and Table 1 in the paper, TLISD outperforms all methods including ILISD by a clear performance margin.
\begin{figure*}[!t]
\centering
\includegraphics[width=\linewidth]{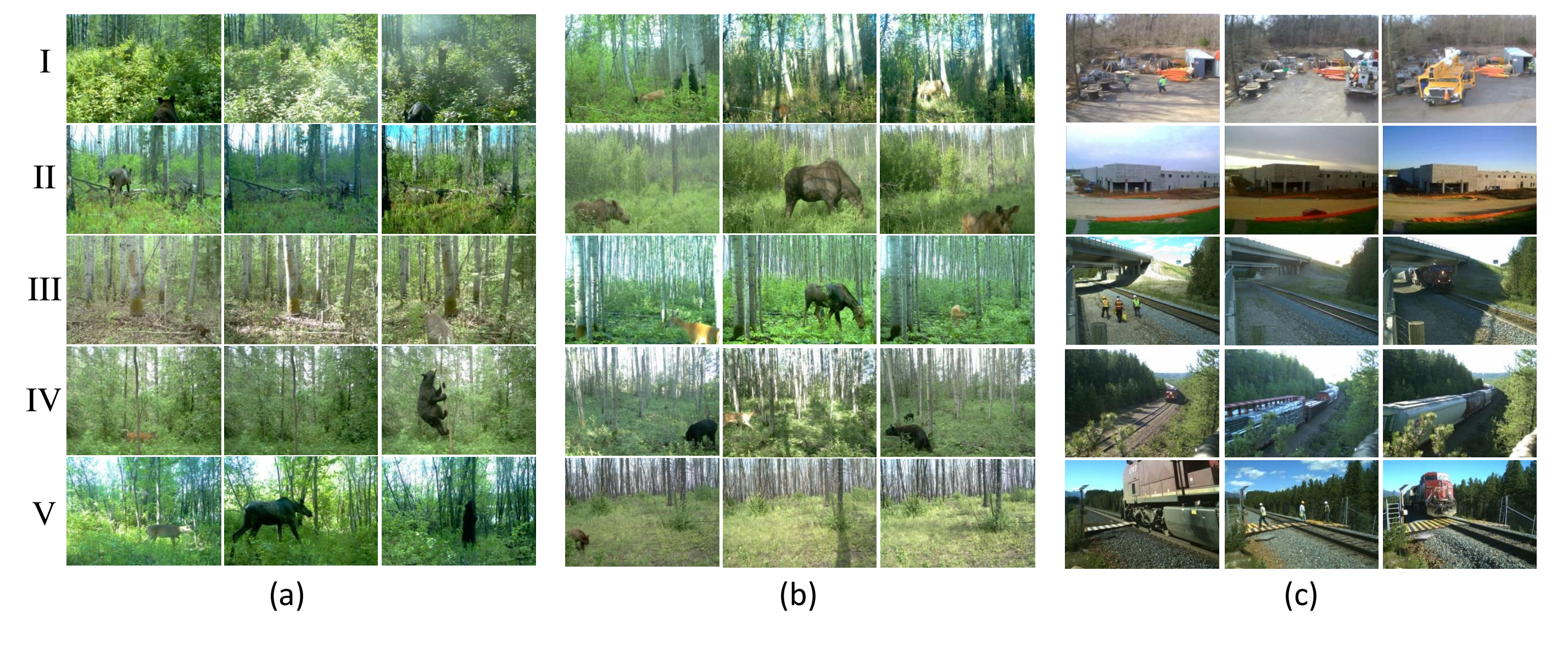}
\caption{Three selected images from each sequence of EIC dataset, captured via surveillance systems in wildlife and industrial applications. Rows in (a) and (b) show 10 wildlife sequences. Rows in (c) show 5 sequences from industrial applications including construction sites and railways sequences}
\label{all_sequence}
\end{figure*}
\subsection{More experimental results on our introduced EIC dataset}
\label{exp_EIC}
As discussed in Section 4.1 in the paper, due to the lack of a comprehensive dataset with various illumination and shadow changes in a real environment, we have created a new benchmark dataset called EIC with around 80k images in 15 sequences, captured via available surveillance systems in wildlife and industrial applications. Fig.~\ref{all_sequence} shows sample images of all EIC sequences. To appreciate the significant variations of illumination we show three images from each sequence. Each row of Figs.~\ref{all_sequence}(a), (b), and (c) shows one of these sequences. Table~\ref{detail} shows the name and image size of each sequence.
 \setlength{\tabcolsep}{4pt}
\begin{table}[t]
\begin{center}
 \resizebox{0.7\linewidth}{!}{
  \begin{tabular}{|l|l|l|}
  \hline
       & Sequence & Image size\\
    \hline
    Fig.~\ref{all_sequence}(a) & I. Wildlife4 & [358,508]\\
&    II. Wildlife5 & [358,508]\\
&    III. Wildlife6 & [358,508]\\
&    IV. Wildlife7 & [358,508]\\
&    V.  Wildlife8 & [358,508]\\
    \hline
   Fig.~\ref{all_sequence}(b)  &  I. Wildlife9 & [358,508]\\
&    II. Wildlife10 & [358,508]\\
&    III. Wildlife11 & [358,508]\\
&    IV. Wildlife12 & [358,508]\\
&    V. Wildlife13 & [358,508]\\
    \hline
   Fig.~\ref{all_sequence}(c)  &  I. Industrial area1 & [350,450]\\
&    II. Industrial area2 & [350,450]\\
&    III. Railway1 & [350,450]\\
&    IV. Railway2 & [350,450]\\
&    V. Railway3 & [350,450]\\
    \hline
  \end{tabular}}
\end{center}
\caption{Name and image size of EIC sequences correspond to the rows in Fig.~\ref{all_sequence}}
\label{detail}
\end{table}
\setlength{\tabcolsep}{1.4pt}

\begin{figure}[!t]
\vspace{-5pt}
\centering
\includegraphics[width=0.9\linewidth, height=11cm]{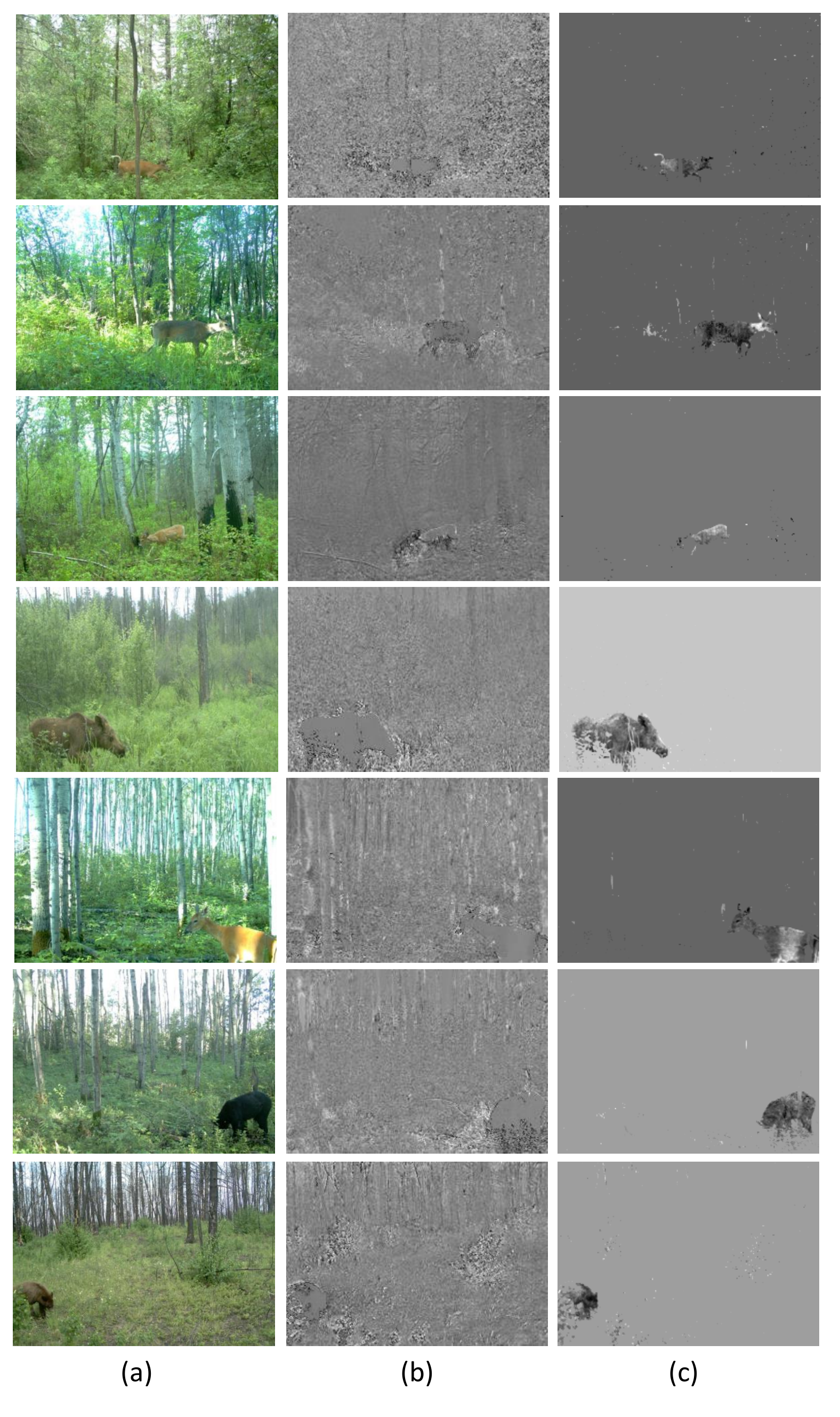}
\caption{Qualitative results of our method (TLISD) on seven wildlife sequences captured by a motion-triggered camera. (a) sample image (b) corresponding illumination changes (c) detected moving objects.}
\label{qualitative_EIC}
\vspace{-25pt}
\end{figure}

\begin{figure*}[!t]
\centering
\includegraphics[width=\linewidth]{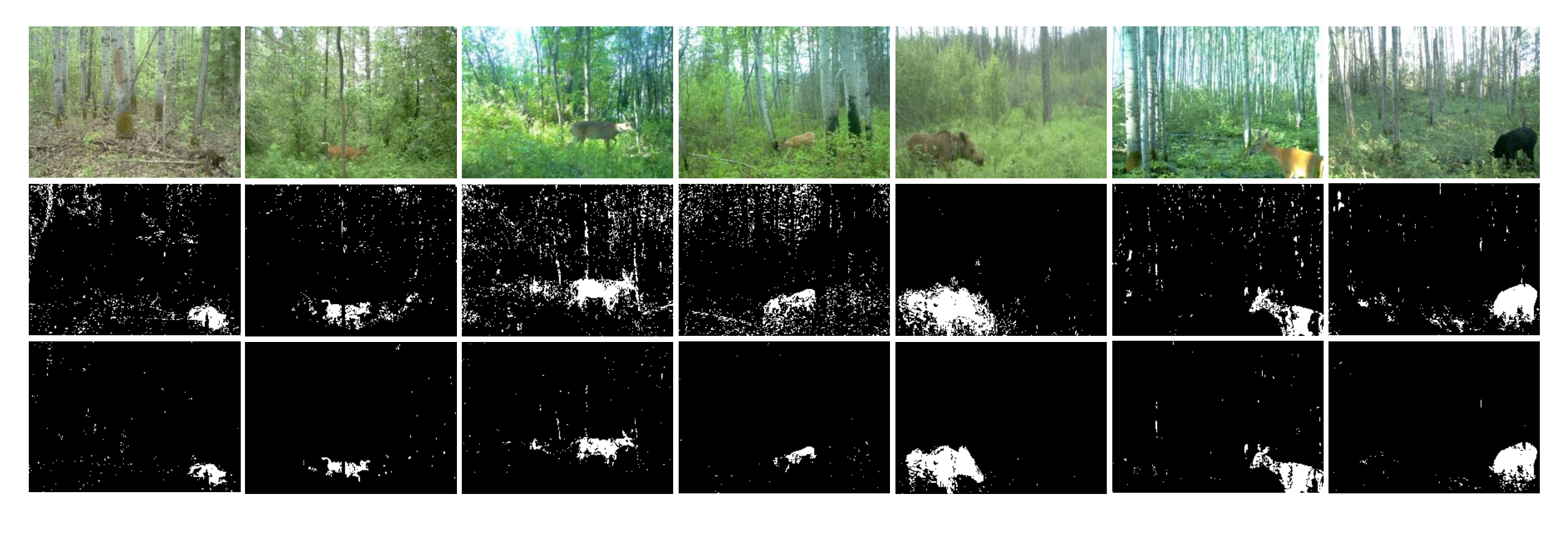}
\caption{Comparison of qualitative results between our method (TLISD) and ILISD on seven sequences of wildlife6, wildlife7, Wildlife8, Wildlife9, Wildlife10, Wildlife11, and Wildlife12. The second row: Qualitative results of ILISD, The third row: Qualitative results of our method (TLISD).}
\label{tlisd_ilisd}
\end{figure*}

\setlength{\tabcolsep}{4pt}
\begin{table*}[!t]
\begin{center}
 \resizebox{0.7\linewidth}{!}{
  \begin{tabular}{|l|c|c|c|c|c|c|c|}
  \hline
    Sequence & Wildlife6 & Wildlife7 & Wildlife8 & Wildlife9 & Wildlife10 & Wildlife11 & Wildlife12\\
    \hline
    ILISD    & 0.6170    & 0.5901    & 0.4836    & 0.5597    & 0.6930     & 0.6852     & 0.6915 \\
    TLISD    & \textbf{0.7522}    & \textbf{0.7706}    & \textbf{0.7022}    & \textbf{0.8061}    & \textbf{0.7898}     & \textbf{0.7518}     & \textbf{0.8114} \\
    \hline
  \end{tabular}}
\end{center}
\caption{Comparison of F-measure score between our proposed method and ILISD on EIC sequences of Fig.~\ref{tlisd_ilisd}}
\label{table_compare}
\end{table*}
\setlength{\tabcolsep}{1.4pt}

To show the capability of TLISD, we examine our method on more sequences of EIC dataset. Since the qualitative and quantitative results of TLISD on six sequences of EIC are shown in Fig.~8 and Table 2 in the paper, here we show qualitative results of the rest of EIC wildlife sequences. Fig.~\ref{qualitative_EIC} shows one sample image from the sequences of ``Wildlife7" to ``Wildlife13" from EIC dataset. The second and the third columns of Fig.~\ref{qualitative_EIC} illustrate the results of our method obtained from the first frontal slice of $\mathcal{C}$, and $\mathcal{S}$, corresponding to illumination changes and moving objects, respectively.

Fig.~\ref{tlisd_ilisd} shows the comparison of qualitative results between TLISD and ILISD. For better comparison between these two methods, we use binary mask on outliers. The second and the third rows show the results of ILISD and TLISD respectively. In all cases, due to use of an inaccurate prior map and the same norm for both illumination and real changes, ILISD generates false positive detections. Since TLISD uses multiple prior maps and two different norms for separating real changes from illumination changes, it can correctly separate those false positive pixels from real changes and can classify them into $\mathcal{C}$ as illumination changes. Table~\ref{table_compare} compares the numerical results of all sequences of Fig.~\ref{tlisd_ilisd} in terms of F-measure, where TLISD outperforms ILISD by a clear performance margin. Both Fig.~\ref{tlisd_ilisd} and Table~\ref{table_compare} show the effect of multiple prior maps and k-support norm in our method to separate illumination changes from real changes qualitatively and quantitatively.

\end{document}